\documentclass[journal]{IEEEtran}

\usepackage{amsmath,amssymb,amsfonts}
\usepackage{graphicx,epsfig,psfrag,xspace}
\usepackage[noend]{algorithmic}
\usepackage[usenames]{color}
\usepackage{subfigure,enumerate}
\usepackage{dblfloatfix}

\graphicspath{{./Figs/}}

\newcommand{\until}[1]{\{1,\dots, #1\}}
\newcommand{\subscr}[2]{#1_{\textup{#2}}}

\newcommand{\setdef}[2]{\{#1 \, | \; #2\}}
\newcommand{\seqdef}[2]{\{#1\}_{#2}}

\newcommand{\switchsig}{\sigma}
\newcommand{\map}[3]{#1: #2 \rightarrow #3}
\newcommand{\setmap}[3]{#1: #2 \rightrightarrows #3}
\newcommand{\union}{\operatorname{\cup}}
\newcommand{\intersection}{\ensuremath{\operatorname{\cap}}}
\newcommand{\intersect}{\ensuremath{\operatorname{\cap}}}

\renewcommand{\theenumi}{(\roman{enumi})}

\newcommand\oprocendsymbol{\hbox{$\square$}}
\newcommand\oprocend{\relax\ifmmode\else\unskip\hfill\fi\oprocendsymbol}

\newtheorem{theorem}{Theorem}[section]
\newtheorem{proposition}[theorem]{Proposition}

\newtheorem{definition}[theorem]{Definition}
\newtheorem{lemma}[theorem]{Lemma}

\newtheorem{remark}[theorem]{Remark}

\newcommand{\qed}{\hfill \mbox{\raggedright \rule{.07in}{.1in}}}

\def\R{\mathbb{R}} 
\newcommand{\real}{\R}
\newcommand{\realnonnegative}{\R_{\geq 0}}
\newcommand{\realpositive}{\ensuremath{\mathbb{R}}_{>0}}
\newcommand{\integersnonnegative}{\ensuremath{\mathbb{Z}}_{\ge 0}}
\newcommand{\integernonnegative}{\ensuremath{\mathbb{Z}}_{\ge 0}}

\newcommand{\Prob}{\mathbb{P}} 

\def\argmin{\mathop{\operatorname{argmin}}}
\newcommand{\card}[1]{\left|#1\right|}  
\newcommand{\ConnPart}{\ensuremath{\operatorname{Part}_N(Q)}}    
\newcommand{\G}{G(Q)}
\newcommand{\AdjG}{\mathcal{G}}                     
\newcommand{\E}{\mathcal{E}}                     
\newcommand{\Cd}{\textup{Cd}}        
\newcommand{\Centroids}{\textup{C}}  

\newcommand{\short}[3]{s_{#1,#2}^{#3}}
\newcommand{\bigO}[1]{\ensuremath{ \mathcal{O} ( #1 )}}
\newcommand{\degree}{\ensuremath{^\circ}}
\newcommand{\search}{\mathbb{D}} 

\newcommand{\Hexp}{\subscr{\mathcal{H}}{expected}}        
\newcommand{\Hgeneric}{\subscr{\mathcal{H}}{multicenter}}
\newcommand{\Hone}{\subscr{\mathcal{H}}{one}}

\newcommand{\PseudoDiam}{\mathcal{S}}
\newcommand{\rcomm}{\subscr{r}{comm}}

\def\clap#1{\hbox to 0pt{\hss#1\hss}}

\newcommand{\speed}{v} 	

\renewcommand{\baselinestretch}{1.0}

\title{Discrete Partitioning and Coverage Control\\for Gossiping Robots
  \thanks{This work was supported in part by ARO MURI Award
    W911NF-05-1-0219, NSF grants IIS-0904501 and
    CPS-1035917, and MIUR grant PRIN-20087W5P2. Preliminary and incomplete versions of this work appeared
    in the Proceedings of the 2009 ASME Dynamic Systems and Control
    Conference, Hollywood, California, USA, and in the Proceedings of the
    2010 IEEE Conference on Decision and Control, Atlanta, Georgia, USA.
    } }

\author{Joseph W. Durham\quad%
  \thanks{Joseph W. Durham and Francesco Bullo are with the
    Department of Mechanical Engineering, University of California, Santa
    Barbara, CA, 93106 {\tt\small (joey, bullo)@engineering.ucsb.edu}}%
  \and Ruggero Carli\quad \thanks{ Ruggero Carli is with the
    Department of Information Engineering, University of Padova, 
    Via Gradenigo 6/a, 35131 Padova, Italy {\tt\small carlirug@dei.unipd.it}}%
  \and Paolo Frasca\quad \thanks{Paolo Frasca is
    with the Dipartimento di Matematica, Politecnico di Torino, corso Duca degli Abruzzi 24, 10129 Torino, Italy
    {\tt\small paolo.frasca@polito.it}}%
  \and Francesco Bullo
}

\begin{document}
\maketitle

\begin{abstract}
We propose distributed algorithms to automatically deploy a team of mobile robots to partition and provide coverage of a non-convex environment. To handle arbitrary non-convex environments, we represent them as graphs. 
 Our partitioning and coverage algorithm requires only short-range, unreliable pairwise ``gossip'' communication. The algorithm has two components: (1) a motion protocol to ensure that neighboring robots communicate at least sporadically, and (2) a pairwise partitioning rule to update territory ownership when two robots communicate. By studying an appropriate dynamical system on the space of partitions of the graph vertices, we prove that territory ownership converges to a pairwise-optimal partition in finite time.  This new equilibrium set represents improved performance over common Lloyd-type algorithms. Additionally, we detail how our algorithm scales well for large teams in large environments and how the computation can run in anytime with limited resources.  Finally, we report on large-scale simulations in complex environments and hardware experiments using the Player/Stage robot control system.
\end{abstract}

\thispagestyle{plain}
\section{Introduction}\label{sec:Intro}

Coordinated networks of mobile robots are already in use for environmental monitoring
and warehouse logistics.
In the near future, autonomous robotic teams will revolutionize transportation of passengers and goods,
search and rescue operations, and other applications.
These tasks share a common feature: the robots are asked to provide service
over a space.  
One question which arises is: when a group of robots is waiting for a task request to come in, how can they best position themselves to be ready to respond?

\begin{figure}[t]
\centering
\includegraphics[width=0.99\columnwidth]{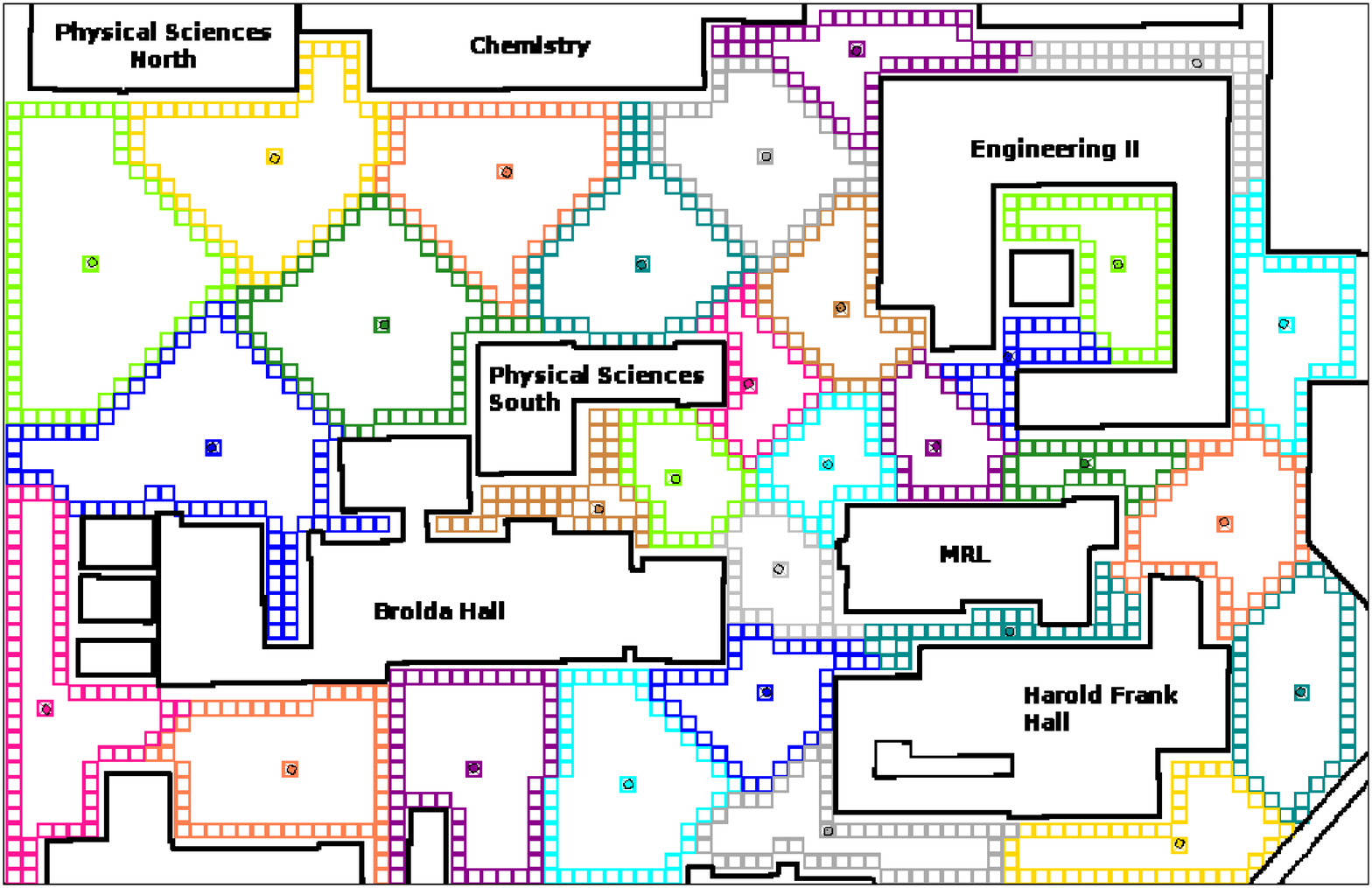}
\caption{Example of a team of robots providing efficient coverage of a non-convex environment, as measured by an appropriate multicenter cost function.}
\label{fig:cover_example}
\end{figure}

The distributed {\em environment partitioning problem} for robotic
networks consists of designing individual control and communication laws such
that the team divides a large space into regions.  Typically, partitioning is
done so as to optimize a cost function which measures the quality of service
provided over all of the regions.  {\em Coverage control} additionally optimizes the positioning
of robots inside a region as shown in Fig.~\ref{fig:cover_example}.

This paper describes a distributed partitioning and coverage control algorithm for a network of robots
to minimize the expected distance between the closest robot and spatially distributed events which will appear at discrete points in a non-convex environment.
Optimality is defined with reference to a relevant ``multicenter'' cost
function.  As with all multirobot coordination
applications, the challenge comes from reducing the communication
requirements: the proposed algorithm requires only short-range ``gossip"
communication, i.e., asynchronous and unreliable 
communication between nearby robots.

\subsection*{Literature Review}

Territory partitioning and coverage control have applications in many fields.
In cyber-physical systems, applications include automated environmental monitoring~\cite{RS-JD-GS:10}, fetching and delivery~\cite{PRW-RdA-MM:08}, construction~\cite{SY-MS-DR:09}, and other vehicle routing scenarios~\cite{FB-EF-MP-KS-SLS:10k}.
More generally, coverage of discrete sets is also closely related to the literature on data clustering and $k$-means~\cite{AKJ-MNM-PJF:99}, as well as the facility location or $k$-center problem~\cite{VVVa:01}.  
Partitioning of graphs is its own field of research, see \cite{POF:98} for a survey.
Territory partitioning through local interactions is also studied for animal groups, see for example~\cite{FRA-DMG:03}.

A broad discussion of algorithms for partitioning and coverage control in robotic networks is presented in~\cite{FB-JC-SM:09} which builds on the classic work of Lloyd~\cite{SPL:82} on optimal quantizer selection through ``centering and partitioning."
The Lloyd approach was first adapted for distributed coverage control in~\cite{JC-SM-TK-FB:02j}.
Since this beginning, similar algorithms have been applied to non-convex environments~\cite{MZ-CGC:08,LCAP-VK-RCM-GASP:08}, unknown density functions~\cite{MS-DR-JJS:08,RC-HT-RL:09}, equitable partitioning~\cite{OB-OB-DK-QW:07}, and construction of truss-like objects~\cite{SY-MS-DR:09}.
There are also multi-agent partitioning algorithms built on market principles or auctions, see~\cite{MBD-RZ-NK-AS:06} for a survey. 
%

While Lloyd iterative optimization algorithms are popular and work well in simulation,
they require synchronous and reliable communication among neighboring robots.  As robots with adjacent regions may be arbitrarily far apart, these communication requirements are burdensome and unrealistic for deployed robotic networks. 
%
In response to this issue, in~\cite{FB-RC-PF:08u-web} the authors have shown how a group of robotic agents can optimize the partition of a convex bounded set using a Lloyd algorithm with gossip communication. 
A Lloyd algorithm with gossip communication has also been applied to optimizing partitions of non-convex environments in~\cite{JWD-RC-PF-FB:08z}, the key idea being to transform the coverage problem in Euclidean space into a coverage problem on a graph with geodesic distances.

Distributed Lloyd methods are built around separate partitioning and centering steps, and they are attractive because there are known ways to characterize their equilibrium sets (the so-called centroidal Voronoi partitions) and prove convergence. Unfortunately, even for very simple environments (both continuous and discrete) the set of centroidal Voronoi partitions may contain several sub-optimal configurations.
We are thus interested in studying (discrete) gossip coverage algorithms for two reasons: (1) they apply to more realistic robot network models featuring very limited communication in large non-convex environments, and (2) they are more flexible than typical Lloyd algorithms meaning they can avoid poor suboptimal configurations and improve performance.

\subsection*{Statement of Contributions}

There are three main contributions in this paper.  
First, we present a discrete partitioning and coverage optimization algorithm for mobile robots with unreliable, asynchronous, and short-range communication.
Our algorithm has two components: a \emph{motion protocol} which drives the robots to meet their neighbors, and a \emph{pairwise partitioning rule} to update territories when two robots meet.
The partitioning rule optimizes coverage of a set of points connected by edges to form a graph.  The flexibility of graphs allows the algorithm to operate in non-convex, non-polygonal environments with holes.  Our graph partition optimization approach can also be applied to non-planar problems, existing transportation or logistics networks, or more general data sets.

Second, we provide an analysis of both the convergence properties and computational requirements of the algorithm.  By studying a dynamical system of partitions of the graph's vertices, we prove that almost surely the algorithm converges to a pairwise-optimal partition in finite time.  The set of pairwise-optimal partitions is shown to be a proper subset of the well-studied set of centroidal Voronoi partitions.
We further describe how our pairwise partitioning rule can be implemented to run in anytime and how the computational requirements of the algorithm can scale up for large domains and large teams.

Third, we detail experimental results from our implementation of the algorithm in the Player/Stage robot control system.  We present a simulation of 30 robots providing coverage of a portion of a college campus to demonstrate that our algorithm can handle large robot teams, and a hardware-in-the-loop experiment conducted in our lab which incorporates sensor noise and uncertainty in robot position.  Through numerical analysis we also show how our new approach to partitioning represents a significant performance improvement over both common Lloyd-type methods and the recent results in~\cite{FB-RC-PF:08u-web}.

The present work differs from the gossip Lloyd method~\cite{FB-RC-PF:08u-web} in three respects.
First, while~\cite{FB-RC-PF:08u-web} focuses on territory partitioning in a convex continuous domain, here we operate on a graph which allows our approach to consider geodesic distances, work in non-convex environments, and maintain connected territories.
Second, instead of a pairwise Lloyd-like update,  we use an iterative optimal two-partitioning approach which yields better final solutions. Third, we also present a motion protocol to produce the sporadic pairwise communications required for our gossip algorithm and characterize the computational complexity of our proposal.
Preliminary versions of this paper appeared in~\cite{JWD-RC-PF-FB:08z} and \cite{JWD-RC-FB:10h}.  Compared to these, the new content here includes: (1) a motion protocol; (2) a simplified and improved pairwise partitioning rule; (3) proofs of the convergence results; and (4) a description of our implementation and a hardware-in-the-loop experiment.

\subsection*{Paper Structure and Notation}

In Section~\ref{sec:prelim} we review and adapt coverage and geometric concepts (e.g., centroids, Voronoi partitions) to a discrete environment like a graph.  We formally describe our robot network model and the discrete partitioning problem in Section~\ref{sec:algorithm}, and then state our coverage algorithm and its properties.
Section~\ref{sec:convergence} contains proofs of the main convergence results.  In Section~\ref{sec:results} we detail our implementation of the algorithm and present experiments and comparative analysis.  Some conclusions are given in Section~\ref{sec:conclusion}.

In our notation, $\realnonnegative$ denotes the set of non-negative real numbers
and $\integernonnegative$ the set of non-negative integers.  Given a
set $A$, $\card{A}$ denotes the number of elements in $A$.
Given sets $A,B$, their difference is $A\setminus B=\setdef{a\in A}{a\notin B}$.  A set-valued map, denoted by $\setmap{T}{A}{B}$, associates to an
element of $A$ a subset of $B$.

\section{Preliminaries}
\label{sec:prelim}

We are given a team of $N$ robots tasked with providing coverage of a finite set of points in a  non-convex and non-polygonal environment.  In this Section we translate concepts used in coverage of continuous environments to graphs.

\subsection{Non-convex Environment as a Graph}

Let $Q$ be a finite set of points in a continuous environment.  These points
represent locations of interest, and are assumed to be connected by weighted edges.
Let $\G=(Q,E,w)$ be an (undirected) weighted graph with edge set
$E\subset Q\times Q$ and weight map $\map{w}{E}{\realpositive}$; we let
$w_{e}>0$ be the weight of edge $e$.
 We assume that $\G$ is connected and think of the edge weights as
distances between locations.

\begin{remark}[Discretization of an Environment]\label{rem:discretization}
 For the examples in this paper we will use a coarse {\em occupancy grid map}
  as a representation of a continuous environment.  In an occupancy grid~\cite{HM:88}, each grid cell is
  either free space or an obstacle (occupied). To form a weighted graph, 
  each free cell becomes a vertex and free cells are connected with edges 
  if they border each other in the grid.
  Edge weights are the distances between the centers of the cells, i.e., the grid resolution.
  There are many other methods to discretize a space, including triangularization and other approaches 
  from computational geometry~\cite{MdB-MvK-MO-OS:00}, which could also be used.
\end{remark}

In any weighted graph $\G$ there is a standard notion of distance between
vertices defined as follows.  A \emph{path} in $G$ is an ordered sequence of
vertices such that any consecutive pair of vertices is an edge of
$G$.  The \emph{weight of a path} is the sum of the weights of the
edges in the path.  Given vertices $h$ and $k$ in $G$, the
\emph{distance} between $h$ and $k$, denoted $d_G(h,k)$, is the weight of
the lowest weight path between them, or $+\infty$ if there is no path. If $G$ is connected,
then the distance between any two vertices in $G$ is finite.
By convention, $d_G(h,k)=0$ if $h=k$. Note
that $d_G(h,k)=d_G(k,h)$, for any $h,k \in Q$.

\subsection{Partitions of Graphs}

We will be partitioning $Q$ into $N$ connected subsets or
regions which will each be covered by an individual robot.  To do so we need to
define distances on induced subgraphs of $\G$.
Given $I\subset Q$, the \emph{subgraph induced by the
restriction of $G$ to $I$}, denoted by $G\intersect{}I$, is the graph
with vertex set equal to $I$ and edge set containing
all weighted edges of $G$ where both vertices belong to $I$. In other
words, we set
$(Q,E,w)\intersect{}I=(Q\intersect{}I,E\intersect{}(I\times{I}),w|_{I\times{I}})$.
The induced subgraph is a weighted graph with a notion of
distance between vertices: given $h,k\in I$, we write
$
  d_I(h,k) := d_{G\intersect{}I}(h,k).
$
Note that $d_I(h,k)\ge d_G(h,k).$

We define a {\em connected subset of $Q$} as a subset
$A \subset Q$ such that $A\neq\emptyset$ and $G \intersect A$ is
connected.
We can then partition $Q$ into connected subsets as follows.
\begin{definition}[Connected Partitions]
\label{def:ConPartitions}
Given the graph $\G=(Q,E,w),$ we define a {\em connected $N-$partition
of $Q$} as a collection $P=\{P_i\}_{i=1}^{N}$ of $N$ subsets of $Q$ such that
\begin{enumerate}
\item ${\bigcup_{i=1}^{N}P_i=Q}$;
\item $P_i\cap P_j=\emptyset$ if $i\neq j$;
\item $P_i\neq\emptyset$ for all $i\in \until{N}$; and
\item $P_i$ is connected for all $i\in \until{N}$. 
\end{enumerate}
Let $\ConnPart$ to be the set of connected $N-$partitions
of $Q$.
\end{definition}

Property (ii) implies that each element of $Q$ belongs to just one $P_i$, i.e., each location in the environment is covered by just one robot.  Notice that each $P_i \in P$ induces a connected subgraph in $\G$.  
In subsequent references to $P_i$ we will often mean $G \intersect{} P_i$, and in fact we refer to $P_i(t)$ as the \emph{dominance subgraph} or \emph{region} of the $i$-th robot at time $t$.  

Among the ways of partitioning $Q$, there are some which are worth special attention. Given a vector of distinct points $c\in Q^N$, the partition $P \in \ConnPart$ is said to be a \emph{Voronoi partition of Q generated by c} if, for each $P_i$ and all $k \in P_i$, we have
$c_i\in P_i$
and
$d_G(k,c_i) \le d_G(k,c_j)$, $\forall j\neq i$.
Note that the Voronoi partition generated by $c$ is not unique since how to apportion tied vertices is unspecified.

\subsection{Adjacency of Partitions}

For our gossip algorithms we need to introduce the notion of adjacent subgraphs. Two distinct connected subgraphs $P_i$, $P_j$ are said to be \emph{adjacent} if there are two vertices $q_i$, $q_j$ belonging, respectively, to $P_i$ and $P_j$ such that $(q_i, q_j) \in E$. Observe that if $P_i$ and $P_j$ are adjacent then $P_i \union P_j$ is connected.
Similarly, we say that robots $i$ and $j$ are adjacent or are neighbors if their subgraphs $P_i$ and $P_j$ are adjacent.
Accordingly, we introduce the following useful notion.
\begin{definition}[Adjacency Graph]
  For $P\in \ConnPart$, we define
  the {\em adjacency graph} between regions of partition $P$ as
  $\AdjG(P)=(\{1,\ldots,N\},\E(P))$, where $(i,j)\in \E(P)$ if $P_i$ and $P_j$ are adjacent.
\end{definition}
Note that $\AdjG(P)$ is always connected since $\G$ is.

\subsection{Cost Functions}

We define three coverage cost functions for graphs: $\Hone$, $\Hgeneric$, and $\Hexp$.
Let the \emph{weight function} $\map{\phi}{Q}{\realpositive}$ assign a relative weight to each element of $Q$.
The {\em one-center function} $\Hone$ gives the cost for a robot to cover a connected subset $A \subset Q$ from a vertex $h\in A$ with relative prioritization set by $\phi$: 
$$\Hone(h; A)={\sum_{k\in A} {d_{A}(h,k)\phi(k)}}.$$
A technical assumption is needed to solve the problem of minimizing $\Hone(\cdot, A)$:  we assume from now on that a {\em total order} relation, $<$, is defined on $Q$, i.e., that $Q=\until{\card{Q}}$.  With this assumption we can deterministically pick a vertex in $A$ which minimizes $\Hone$ as follows.

\begin{definition}[Centroid]\label{def:Centroid}
Let $Q$ be a totally ordered set, and let $A \subset Q$. We define the set of generalized centroids of $A$ as the set of vertices in $A$ which minimize $\Hone$, i.e.,
\begin{align*}
\Centroids(A):=\argmin_{h\in A} \Hone(h;A).
\end{align*}
Further, we define the map $\Cd$ as $\Cd(A) := \min\{ c\in \Centroids(A) \}$.  We call $\Cd(A)$ the \emph{generalized centroid} of $A$.
\end{definition}

In subsequent use we drop the word ``generalized" for brevity. Note that with this definition the centroid is well-defined, and also that the centroid of a region always belongs to the region.
With a slight notational abuse, we define $\map{\Cd}{\ConnPart}{Q^N}$ as the map which associates to a partition the vector of the centroids of its elements.

We define the \emph{multicenter function} 
$\Hgeneric$ to measure the cost for $N$ robots to cover a connected $N$-partition $P$ from the vertex set $c \in Q^N$:
$$\Hgeneric(c,P)=\frac{1}{\sum_{k\in Q} \phi(k)}\sum_{i=1}^{N} \Hone(c_i; P_i).$$
We aim to minimize the performance function $\Hgeneric$ with respect to both the vertices $c$ and the partition $P$.

We can now state the coverage cost function we will be concerned with for the rest of this paper.  Let $\map{\Hexp}{\ConnPart}{\realnonnegative}$
be defined by
\begin{align*}
\Hexp(P) &= \Hgeneric(\Cd(P),P).
\end{align*}
In the motivational scenario we are considering, each robot will periodically be asked to perform a task somewhere in its region with tasks appearing according to distribution $\phi$.  When idle, the robots would position themselves at the centroid of their region.  By partitioning $G$ so as to minimize $\Hexp$, the robot team would minimize the expected distance between a task and the robot which will service it.

\subsection{Optimal Partitions}

We introduce two notions of optimal partitions: centroidal Voronoi and pairwise-optimal.
Our discussion starts with the following simple result about the multicenter cost function.
\begin{proposition}[Properties of Multicenter Function]\label{prop:optimal-for-Hgeneric}
Let $P\in \ConnPart$ and $c\in Q^N$. If $P'$ is a Voronoi partition generated by $c$ and
$c' \in Q^n$ is such that $c'_i \in \Centroids(P_i) ~\forall~ i$, then
\begin{align*}
\Hgeneric(c,P') & \le \Hgeneric(c,P), \text{and}\\
\Hgeneric(c',P) &\le \Hgeneric(c,P).
\end{align*}
The second inequality is strict if any $c_i \notin \Centroids(P_i)$.
\end{proposition}

Proposition~\ref{prop:optimal-for-Hgeneric} implies the following necessary condition:
if $(c,P)$ minimizes $\Hgeneric$, then $c_i \in \Centroids(P_i) ~ \forall i$ and $P$ must be a Voronoi partition generated by~$c$.
Thus, $\Hexp$ has the following property as an immediate consequence of Proposition~\ref{prop:optimal-for-Hgeneric}: given $P\in \ConnPart$, if $P^*$ is a Voronoi partition generated by $\Cd(P)$ then
$$
\Hexp(P^*)\leq \Hexp(P).
$$
This fact motivates the following definition.
\begin{definition}[Centroidal Voronoi Partition]
$P\in \ConnPart$ is a \emph{centroidal Voronoi partition} of $Q$ if there exists a $c\in Q^n$ such that $P$ is a Voronoi partition generated by $c$ and $c_i \in \Centroids(P_i) ~\forall~ i$.
\end{definition}

\smallskip
The set of \emph{pairwise-optimal partitions} provides an alternative definition for the optimality of a partition: a partition is pairwise-optimal if, for every pair of adjacent regions, one can not find a better two-partition of the union of the two regions.
This condition is formally stated as follows.
\begin{definition}[Pairwise-optimal Partition]
$P\in \ConnPart$ is a \emph{pairwise-optimal partition}  if for every $(i,j)\in \E(P)$, 
\begin{align*}
&\quad \Hone(\Cd(P_i); P_i)+\Hone(\Cd(P_j); P_j)=\\
&~\min_{a, b \in P_i \cup P_j} \biggl\{\sum_{k\in P_i \cup P_j} \min \left\{d_{P_i \cup P_j} (a,k), d_{P_i \cup P_j} (b,k)\right\} \phi(k) \biggr\}.
\end{align*}
\end{definition}

\begin{figure}[tbp]
\centering
\subfigure[]
{
    \includegraphics[angle=0,width=0.28\columnwidth]{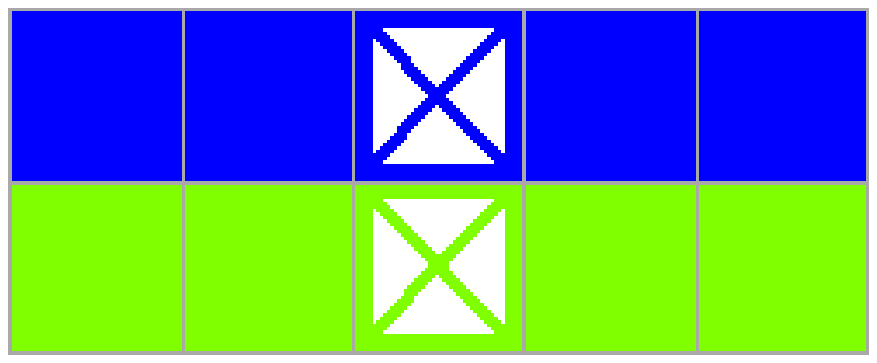}
}
\hfill
\subfigure[]
{
    \includegraphics[angle=0,width=0.28\columnwidth]{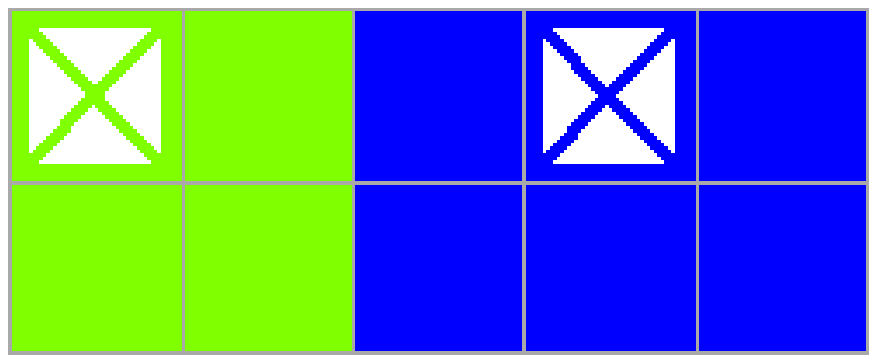}
}
\hfill
\subfigure[]
{
    \includegraphics[angle=0,width=0.28\columnwidth]{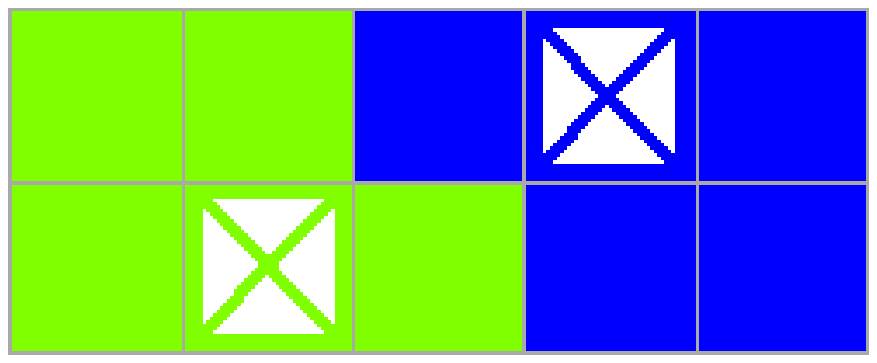}
}

\caption{All possible centroidal Voronoi partitions of a uniform $2 \times 5$ grid.  Assuming all edge weights are $w$ and all vertices have priority $1$, then (a) has a cost of $1.2 w$, (b) has a cost of $1.1 w$, and (c) has a cost of $1.0 w$.  Only (c) is pairwise-optimal by definition.}
\label{fig:voronoi}
\end{figure}

The following Proposition states that the set pairwise-optimal partitions is in fact a subset of the set of centroidal Voronoi partitions.  The proof is involved and is deferred to Appendix~\ref{sec:appendix_C}.  See Fig.~\ref{fig:voronoi} for an example which demonstrates that the inclusion is strict.

\begin{proposition}[Pairwise-optimal Implies Voronoi]\label{prop:OptPair}
Let $P \in \ConnPart$ be a \emph{pairwise-optimal partition}. Then $P$ is also a \emph{centroidal Voronoi partition}.
\end{proposition}

For a given environment $Q$, a pair made of a centroidal Voronoi partition $P$ and the corresponding vector of centroids $c$ is locally optimal in the following sense:  $\Hexp$ cannot be reduced by changing either $P$ or $c$ independently.
A pairwise-optimal partition achieves this property and adds that  
for every pair of neighboring robots $(i,j)$, there does not exist a two-partition of $P_i \cup P_j$ with a lower coverage cost.
In other words, positioning the robots at the centroids of a centroidal Voronoi partition (locally) minimizes the expected distance between a task
appearing randomly in $Q$ according to relative weights 
$\phi$ and the robot who owns the vertex where the task appears.  Positioning at the centroids of a pairwise-optimal partition improves performance by reducing the number of sub-optimal solutions which the team might converge to.

\section{Models, Problem Formulation, and Proposed Solution}
\label{sec:algorithm}

We aim to partition $Q$ among $N$ robotic agents using only asynchronous, unreliable, short-range communication.  
In Section~\ref{sec:Model} we describe the computation, motion, and communication capabilities required of the team of robots, and in Section~\ref{sec:ProblemFormulation} we formally state the problem we are addressing. In Section~\ref{sec:Algorithm} we propose our solution, the \emph{Discrete Gossip Coverage Algorithm}, and in \ref{sec:illustrative} we provide an illustration.  In Sections~\ref{sec:ConvProp} and \ref{sec:computation} we state the algorithm's convergence and complexity properties.

\subsection{Robot Network Model with Gossip Communication}\label{sec:Model}

Our Discrete Gossip Coverage Algorithm requires a team of $N$ robotic agents where each agent $i\in \until{N}$ has the following basic computation and motion capabilities:
\begin{enumerate}
\item [(C1)] agent $i$ knows its unique identifier $i$;
\item [(C2)] agent $i$ has a processor with the ability to store $\G$ and perform operations on subgraphs of $G(Q)$; and 
\item [(C3)] agent $i$ can determine which vertex in $Q$ it occupies and can move at speed $\speed$ along the edges of $\G$ to any other vertex in $Q$.
\end{enumerate}
\begin{remark}[Localization]
The localization requirement in (C3) is actually quite loose.  Localization is only used for navigation and not for updating partitions, thus limited duration localization errors are not a problem.
\end{remark}

The robotic agents are assumed to be able to communicate with each other according to the \emph{range-limited gossip communication model} which is described as follows:
\begin{enumerate}
\item [(C4)]  given a communication range $\rcomm > \max_{e \in E} w_e$, when any two agents reside for some positive duration at a distance $r < \rcomm$, they communicate at the sample times of a Poisson process with intensity $\lambda_{\text{comm}} > 0$.
\end{enumerate}
Recall that an homogeneous Poisson process is a widely-used stochastic model for events which occur randomly and independently in time, where the expected number of events in a period $\Delta$ is $\Delta \lambda_{\text{comm}}$.
%

\begin{remark}[Communication Model]\label{rem:comm}
\noindent (1)~~This communication capability is the minimum necessary for our algorithm, any additional capability can only reduce the time required for convergence.
For example, it would be acceptable to have
intensity $\lambda(r)$ depend upon the pairwise robot distance
in such a way that $\lambda(r) \geq \lambda_{\text{comm}}$ for $r < \rcomm$.

\noindent (2)\quad We use distances in the graph to model limited range communication.  These graph distances are assumed to approximate geodesic distances in the underlying continuous environment and thus path distances for a diffracting wave or moving robot.
\end{remark}

\subsection{Problem Statement}\label{sec:ProblemFormulation}

Assume that, for all $t\in\realnonnegative$, each agent $i\in \until{N}$ maintains in memory a connected subset $P_i(t)$ of environment $Q$.
Our goal is to design a distributed algorithm that iteratively updates the partition $P(t)=\{P_i(t)\}_{i=1}^{N}$ while solving the following optimization problem:
\begin{equation}\label{eq:min}
\min_{P\in \ConnPart} \Hexp(P),
\end{equation}
subject to the constraints imposed by the robot network model with range-limited gossip communication from Section~\ref{sec:Model}.

\begin{figure*}[tbp]
\centering
\subfigure
{
    \includegraphics[width=0.185\textwidth]{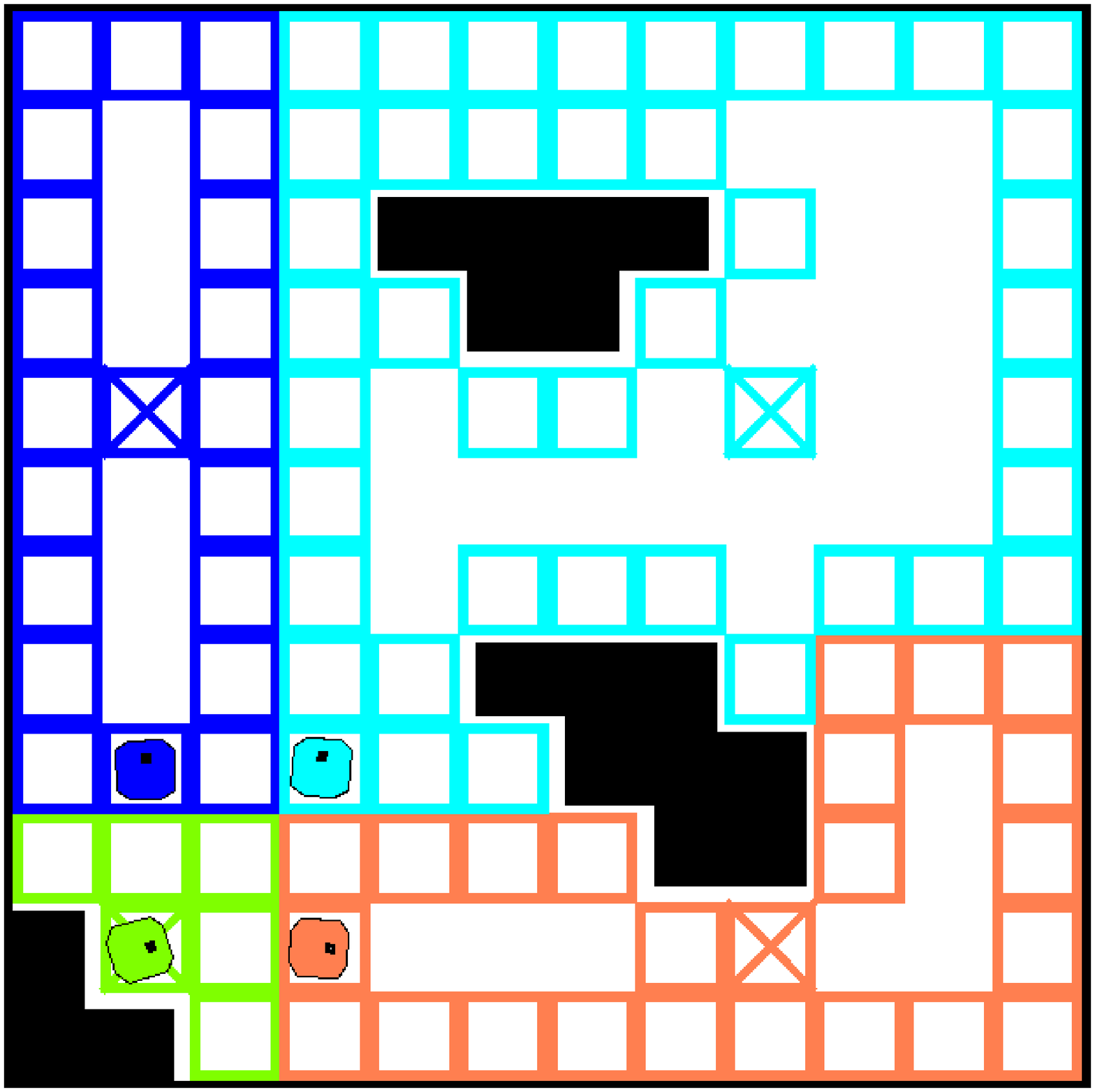}
    \label{fig:sim_four_initial_centroids}
}
\hspace{-0.1in}
\subfigure
{
    \includegraphics[width=0.185\textwidth]{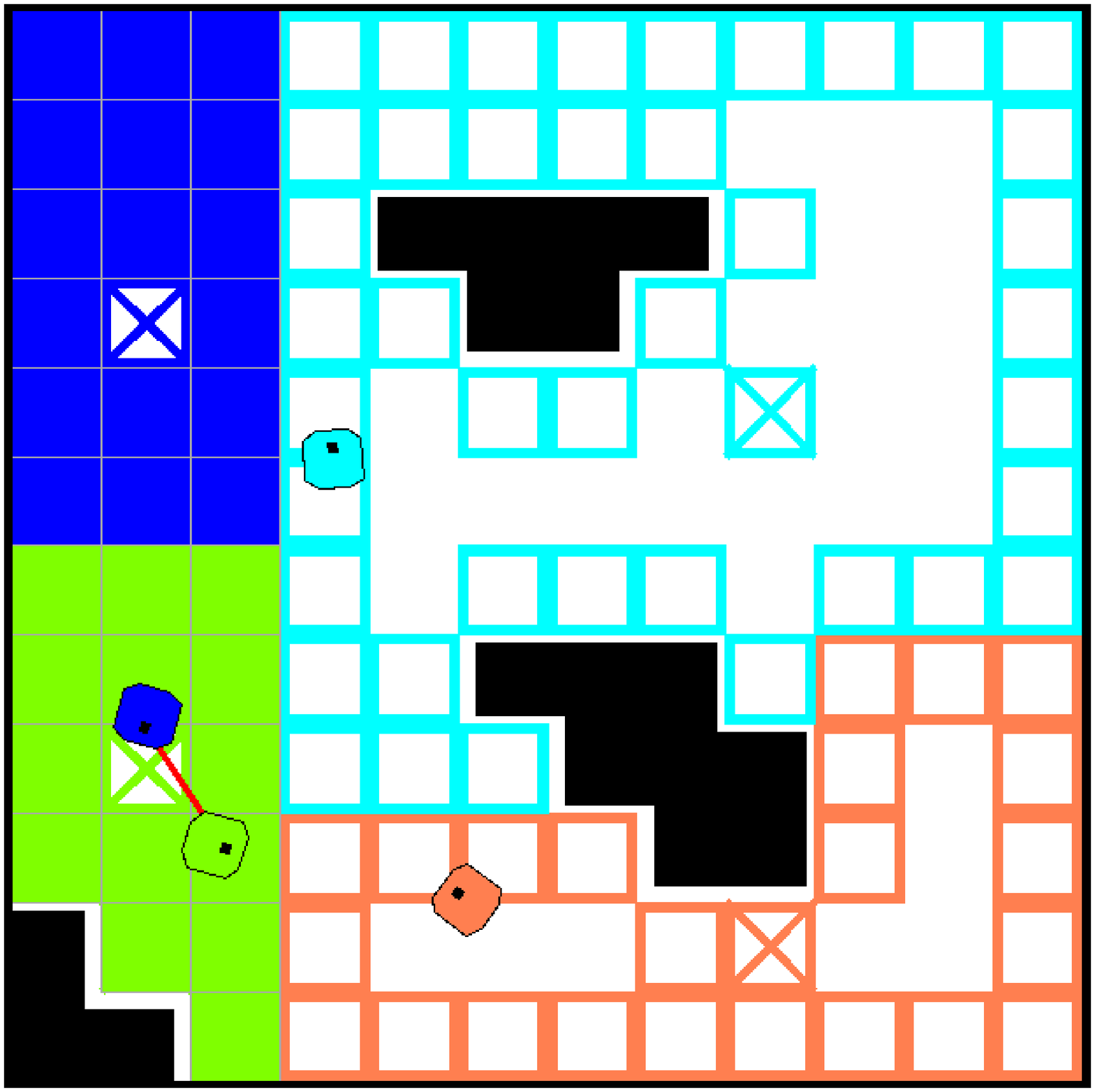}
    \label{fig:sim_four_swap}
}
\hspace{-0.1in}
\subfigure
{
    \includegraphics[width=0.185\textwidth]{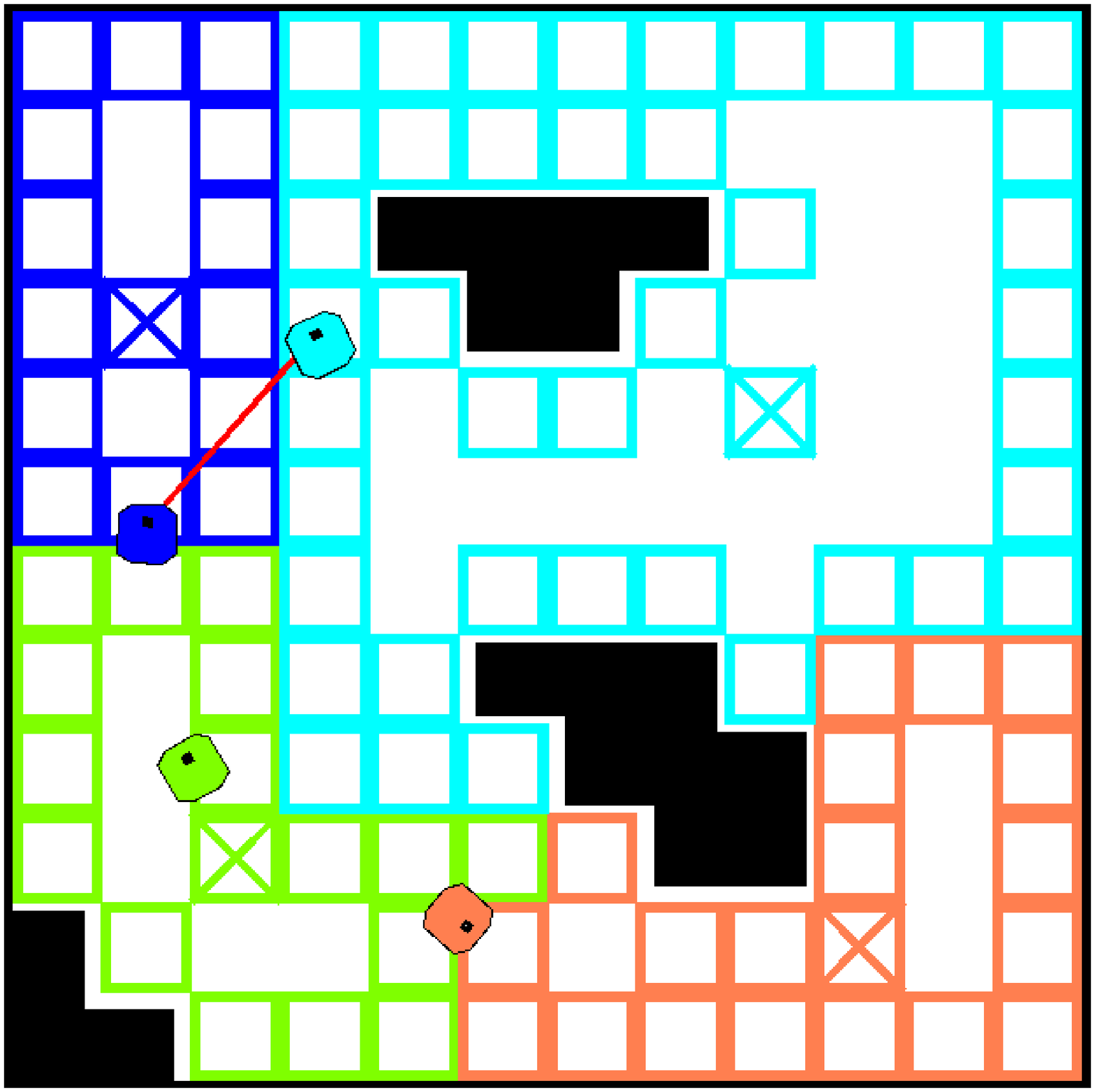}
    \label{fig:sim_four_swap2}
}
\hspace{-0.1in}
\subfigure
{
    \includegraphics[width=0.185\textwidth]{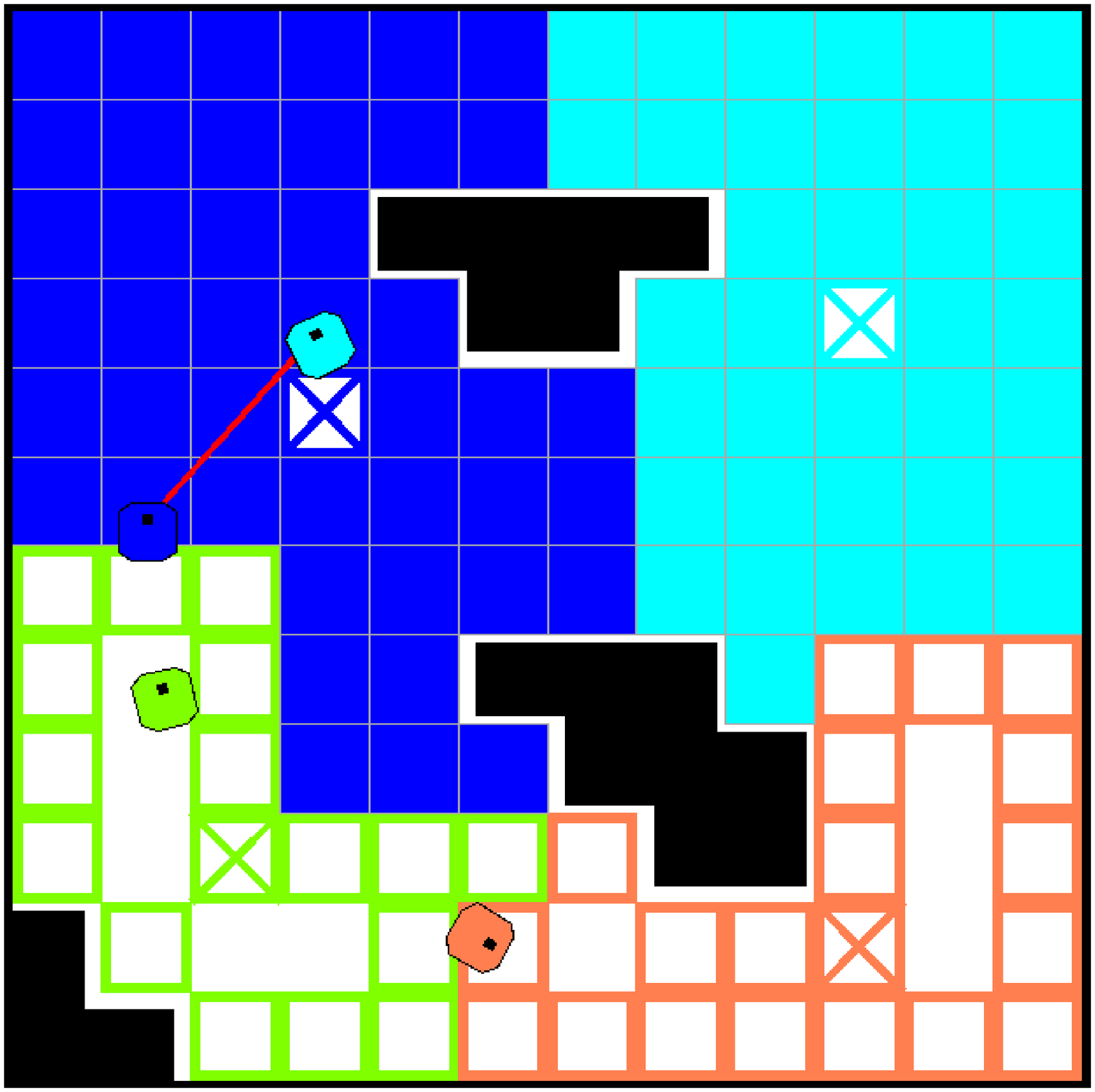}
    \label{fig:sim_four_swap3}
}
\hspace{-0.1in}
\subfigure
{
    \includegraphics[width=0.185\textwidth]{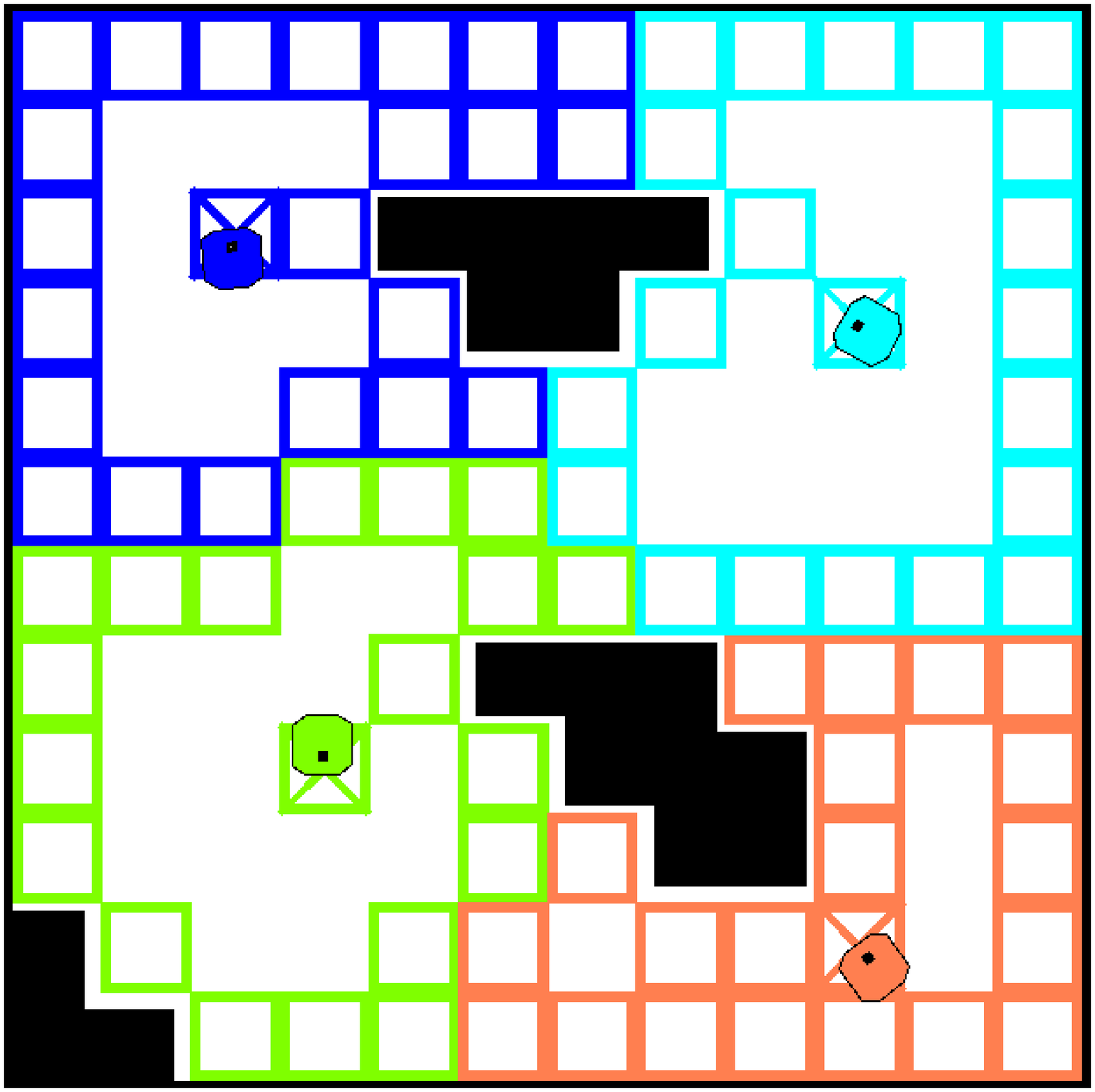}
    \label{fig:sim_four_final}
}
\caption{Simulation of four robots dividing a square environment with obstacles.  The boundary of each robots territory is drawn in a different color, the centroid of a territory is drawn with an X, and pairwise communication is drawn with a solid red line.  On the left is the initial partition assigned to the robots.  The middle frames show two pairwise territory exchanges, with updated territories highlighted with solid colors.  The final partition is shown at right.}
\label{fig:sim_four}
\end{figure*}

\subsection{The Discrete Gossip Coverage Algorithm}\label{sec:Algorithm}

In the design of an algorithm for the minimization problem~\eqref{eq:min}
there are two main questions which must be addressed.
First, given the limited communication capabilities in (C4), how should the robots move inside $Q$
to guarantee frequent enough meetings between pairs of robots?
Second, when two robots are communicating, what information should they exchange and how should they update their regions?

In this section we introduce the \emph{Discrete Gossip Coverage Algorithm} which, following these two questions, consists of two components:
\begin{enumerate}[(1)]
\item the \emph{Random Destination \& Wait Motion Protocol}; and
\item the \emph{Pairwise Partitioning Rule}.
\end{enumerate}
The concurrent implementation of the Random Destination \& Wait Motion Protocol and the Pairwise Partitioning Rule determines the evolution of the positions and dominance subgraphs of the agents as we now formally describe. We start with the Random Destination \& Wait Motion Protocol.


\renewcommand\footnoterule{\hrule width \linewidth height .4pt}

\medskip\footnoterule\vspace{.75\smallskipamount}
\noindent\hfill\textbf{Random Destination \& Wait Motion Protocol}\hfill\vspace{.75\smallskipamount}
\footnoterule\vspace{.75\smallskipamount}
\noindent Each agent $i \in \until{N}$ determines its motion by repeatedly performing the following actions:
\begin{algorithmic}[1]
  \STATE agent $i$ samples a \emph{destination vertex} $q_i$ from a uniform  distribution over its dominance subgraph $P_i$;

  \STATE agent $i$ moves to vertex $q_i$ through the shortest path in $P_i$ connecting the vertex it currently occupies and $q_i$; and

  \STATE agent $i$ waits at $q_i$ for a duration $\tau>0$.
\end{algorithmic}
\vspace{.5\smallskipamount}\footnoterule\smallskip

If agent $i$ is moving from one vertex to another we say that agent $i$ is in the \emph{moving} state while if agent $i$ is waiting at some vertex we say that it is in the \emph{waiting} state. 

\begin{remark}[Motion Protocol]
The motion protocol is designed to ensure frequent enough communication between pairs of robots. In general, any motion protocol can be used which meets this requirement, so $i$ could select $q_i$ from the boundary of $P_i$ or use some heuristic non-uniform distribution over $P_i$. 
\end{remark}

If any two agents $i$ and $j$ reside in two vertices at a graphical distance smaller that $\rcomm$ for some positive duration, then at the sample times of the corresponding communication Poisson process the two agents exchange sufficient information to update their respective dominance subgraphs $P_i$ and $P_j$ via the Pairwise Partitioning Rule.


\renewcommand\footnoterule{\hrule width \linewidth height .4pt}

\medskip\footnoterule\vspace{.75\smallskipamount}
\noindent\hfill\textbf{Pairwise Partitioning Rule}\hfill\vspace{.75\smallskipamount}
\footnoterule\vspace{.75\smallskipamount}

\noindent 
Assume that at time $t\in\realnonnegative$, agent $i$ and agent $j$ communicate. Without loss of generality assume that $i<j$.  Let $P_i(t)$ and $P_j(t)$ denote the current dominance subgraphs of $i$ and $j$, respectively. Moreover, let $t^+$ denote the time instant just after $t$.  Then, agents $i$ and $j$ perform the following tasks:

\begin{algorithmic}[1]
  \STATE agent $i$ transmits $P_i(t)$ to agent $j$ and vice-versa
  \STATE initialize $W_{a^*} := P_i(t)$, $W_{b^*} := P_j(t)$, $a^* := \Cd(P_i(t))$, $b^* := \Cd(P_j(t))$
  \STATE compute $U:=P_i(t)\union P_j(t)$ and an ordered list $S$ of all pairs of vertices in $U$
  \FOR{each $(a,b) \in S$}
	  \STATE compute the sets\\
	      \qquad $W_{a} :=\left\{x\in U : d_{U}(x, a) \leq d_{U}(x, b)\right\}$ \\
	      \qquad $W_{b} :=\left\{x\in U : d_{U}(x, a) > d_{U}(x, b)\right\}$
	  \STATE \textbf{if~} $\Hone(a; W_{a}) ~+~ \Hone(b;W_{b}) < $ \\ 
	      $\quad~ \Hone(a^*; W_{a^*}) + \Hone(b^*;W_{b^*})$ \textbf{~then}
	      \STATE $\qquad W_{a^*}:=W_{a}, W_{b^*}:=W_{b}, a^* := a, b^* := b$
  \ENDFOR
  \STATE $P_i(t^+):=W_{a^*}, \quad P_j(t^+):=W_{b^*}$
\end{algorithmic}
\vspace{.5\smallskipamount}\footnoterule\smallskip

Some remarks are now in order.

\begin{remark}[Partitioning Rule]
\noindent (1)~~ 
The Pairwise Partitioning Rule is designed to find a minimum cost two-partition of $U$. More formally,  if list $S$ and sets $W_{a^*}$ and $W_{b^*}$ for $(a^*,b^*)\in S$ are defined as in the Pairwise Partitioning Rule, then $W_{a^*}$ and $W_{b^*}$ are an optimal two-partition of $U$.

\noindent (2)\quad 
While the loop in steps 4-7 must run to completion to guarantee that $W_{a^*}$ and $W_{b^*}$ are an optimal two-partition of $U$, the loop is designed to return an intermediate sub-optimal result if need be.  If $P_i$ and $P_j$ change, then $\Hexp$ will decrease and this is enough to ensure eventual convergence.

\noindent (3)\quad  We make a simplifying assumption in the Pairwise Partitioning Rule that, once two agents communicate, the application of the partitioning rule is instantaneous.  We discuss the actual computation time required in Section~\ref{sec:computation} and some implementation details in Section~\ref{sec:results}.

\noindent (4)\quad  Notice that simply assigning $W_{a^*}$ to $i$ and $W_{b^*}$ to $j$ can cause the robots to ``switch sides'' in $U$.  While convergence is guaranteed regardless, switching may be undesirable in some applications. In that case, any smart matching of $W_{a^*}$ and $W_{b^*}$ to $i$ and $j$ may be inserted.

\noindent (5)\quad Agents who are not adjacent may communicate but the partitioning rule will not change their regions. Indeed, in this case $W_{a^*}$ and $W_{b^*}$ will not change from $P_i(t)$ and $P_j(t)$.
\end{remark}

Some possible modifications and extensions to the algorithm are worth mentioning.
\begin{remark}[Heterogeneous Robotic Networks]
In case the robots have heterogeneous dynamics, line 5 can be modified to consider per-robot travel times between vertices. For example, $d_U(x,a)$ could be replaced by the expected time for robot $i$ to travel from $a$ to $x$ while $d_U(x,b)$ would consider robot $j$.
\end{remark}
\begin{remark}[Coverage and Task Servicing]
Here we focus on partitioning territory, but this algorithm can easily be combined with methods to provide a service in $Q$ as in~\cite{FB-EF-MP-KS-SLS:10k}.  The agents could split their time between moving to meet their neighbors and update territory, and performing requested tasks in their region.
\end{remark}

\subsection{Illustrative Simulation}
\label{sec:illustrative}

The simulation in Fig.~\ref{fig:sim_four} shows four
robots partitioning a square environment with obstacles where the free space is represented by a $12 \times 12$ grid.  In the initial partition shown in the left panel,  the robot in the top
right controls most of the environment while the robot in the bottom
left controls very little.  The robots then move according to 
the Random Destination \& Wait Motion Protocol, and communicate
according to range-limited gossip communication model with $r_{comm} = 2.5m$
(four edges in the graph).

The first pairwise territory exchange is shown in the second panel, where the bottom left robot claims some territory from the robot on the top left.
A later exchange between the two robots on the top is shown in the next two panels.  Notice that the cyan robot in the top right gives away the vertex it currently occupies.  In such a scenario, we direct the robot to follow the shortest path in $\G$ to its updated territory before continuing on to a random destination.

After 9 pairwise territory exchanges, the robots reach the pairwise-optimal partition shown at right in Fig.~\ref{fig:sim_four}.
The expected distance between a random vertex and the closest robot decreases from $2.34m$ down to $1.74m$.

\subsection{Convergence Property}
\label{sec:ConvProp}

The strength of the Discrete Gossip Coverage Algorithm is the possibility of enforcing that a
partition will converge to a pairwise-optimal partition through pairwise territory exchange.
In Theorem~\ref{th:main} we summarize this convergence property, with proofs given in Section~\ref{sec:convergence}. 

\begin{theorem}[Convergence Property]\label{th:main}
Consider a network of $N$ robotic agents endowed with computation and motion capacities (C1), (C2), (C3), and communication capacities (C4). Assume the agents implement the \emph{Discrete Gossip Coverage Algorithm} consisting of the concurrent implementation of the \emph{Random Destination \& Wait Motion Protocol} and the \emph{Pairwise Partitioning Rule}. Then, 
\begin{enumerate}[(i)]
\item\label{item:well-posedness} the partition $P(t)$ remains connected and is described by $P:\realnonnegative \to \ConnPart,$ and
\item\label{item:convergence} $P(t)$ converges almost surely in finite time to a pairwise-optimal partition.
\end{enumerate} 
\end{theorem}

\begin{remark}[Optimality of Solutions]
By definition, a pairwise-optimal partition is optimal in that $\Hexp$ can not be improved by changing only two regions in the partition.
\end{remark}

\begin{remark}[Generalizations]
For simplicity we assume uniform robot speeds, communication processes, and waiting times.  An extension to non-uniform processes would be straightforward.
\end{remark}

\subsection{Complexity Properties and Discussion}
\label{sec:computation}

In this subsection we explore the computational requirements
of the Discrete Gossip Coverage Algorithm, and make some comments on implementation.  
Cost function $\Hone(h; P_i)$ is the sum of the distances between $h$ and all other vertices in $P_i$.  This computation of one-to-all distances is the core computation of the
algorithm.  
For most graphs of interest the total number of edges $\card{E}$ is proportional to $\card{Q}$, so we will state bounds on this computation in terms of $\card{P_i}$.
Computing one-to-all distances requires one of the following:
\begin{itemize}
\item if all edge weights in $\G$ are the same (e.g., for a graph from an occupancy grid), a breadth-first search approach can be used which requires \bigO{\card{P_i}} in time
and memory;
\item otherwise, Dijkstra's algorithm must be used which requires \bigO{\card{P_i} \log\left(\card{P_i}\right)} in time and \bigO{\card{P_i}} in memory.
\end{itemize}
Let $\search(P_i)$ be the time to compute one-to-all distances in $P_i$, then computing $\Hone(h; P_i)$ requires \bigO{\search(P_i)} in time.

\begin{proposition}[Complexity Properties]
\label{prop:computation}
The motion protocol requires 
\bigO{\card{P_i}} in memory, and \bigO{\search(P_i)} in computation time.
The partitioning rule requires
\bigO{\card{P_i} + \card{P_j}} in communication bandwidth between robots $i$ and $j$, \bigO{\card{P_i} + \card{P_j}} in memory, and can run in any time.
\end{proposition}
\begin{IEEEproof}
We first prove the claims for the motion protocol.  Step 2 is the only non-trivial step and requires finding a shortest path in $P_i$, which is equivalent to computing one-to-all distances from the robot's current vertex.  Hence, it requires \bigO{\search(P_i)} in time and \bigO{P_i} in memory.

We now prove the claims for the partitioning rule.
In step 1, robots $i$ and $j$ transmit their subgraphs to each other, which requires \bigO{\card{P_i} + \card{P_j}} in communication bandwidth.  For step 3, the robots determine $U := P_i \cup P_j$, which requires \bigO{\card{P_i} + \card{P_j}} in memory to store.  
Step 4 is the start of a loop which executes \bigO{\card{U}^2} times, affecting the time complexity of steps 5, 6 and 7.
Step 5 requires two computations of one-to-all distances in $U$ which each take \bigO{\search(U)}.
Step 6 involves four computations of $\Hone$ over different subsets of $U$, however those for $W_{a^*}$ and $W_{b^*}$ can be stored from previous computation.  Since $W_a$ and $W_b$ are strict subsets of $U$, step 5 takes longer than step 6.
Step 7 is trivial, as is step 8.
The total time complexity of the loop is thus \bigO{\card{U}^2 \search(U)}.

However, the loop in steps 4-7 can be truncated after any number of iterations.  While it must run to completion to guarantee that $W_{a^*}$ and $W_{b^*}$ are an optimal two-partition of $U$, the loop is designed to return an intermediate sub-optimal result if need be.  If $P_i$ and $P_j$ change, then $\Hexp$ will decrease.  Our convergence result will hold provided that all elements of $S$ are eventually checked if $P_i$ and $P_j$ do not change.
Thus, the partitioning rule can run in any time with each iteration requiring \bigO{\search(U)}.
\end{IEEEproof}

All of the computation and communication requirements in Proposition~\ref{prop:computation} are independent of the number of robots and scale with the size of a robot's partition, meaning the Discrete Gossip Coverage Algorithm can easily scale up for large teams of robots in large environments.

\section{Convergence Proofs}
\label{sec:convergence}

This section is devoted to proving the two statements in Theorem~\ref{th:main}.  
The proof that the Pairwise Partitioning Rule maps a connected $N$-partition into a connected $N$-partition is straightforward. The proof of convergence is more involved and is based on the application of Lemma~\ref{lem:finite-LaSalle} in Appendix~\ref{sec:appendix_A} to the Discrete Gossip Coverage Algorithm. Lemma~\ref{lem:finite-LaSalle} establishes strong convergence properties for a particular class of set valued maps (set-valued maps are briefly reviewed in Appendix~\ref{sec:appendix_A}).

We start by proving that the Pairwise Partitioning Rule is well-posed in the sense that it maintains a connected partition.


\begin{IEEEproof}[Proof of Theorem~\ref{th:main} statement (\ref{item:well-posedness})]
To prove the statement we need to show that $P(t^+)$ satisfies points (i) through (iv) of Definition~\ref{def:ConPartitions}. From the definition of the Pairwise Partitioning Rule, we have that $P_i(t^+) \cup P_j(t^+)=P_i(t) \cup P_j(t)$ and $P_i(t^+) \cap P_j(t^+) = \emptyset$. Moreover, since $a^*\in P_i(t^+)$ and $b^*\in P_j(t^+)$, it follows that $P_i(t^+)\neq \emptyset$ and $P_j(t^+)\neq \emptyset$. These observations imply the validity of points (i), (ii), and (iii) for $P(t^+)$. Finally, 
we must show that $P_i(t^+)$ and $P_j(t^+)$ are connected, i.e., $P(t^+)$ also satisfies point (iv).
To do so we show that, given $x\in W_{a^*}$, any shortest path in $P_i(t) \cup P_j(t)$ connecting $x$ to $a^*$ completely belongs to $W_{a^*}$. We proceed by contradiction. Let $s_{x,a^*}$ denote a shortest path in $P_i(t)\cup P_j(t)$ connecting $x$ to $a^*$ and let us assume that there exists $m\in s_{x,a^*}$ such that $m\in W_{b^*}$. For $m$ to be in $W_{b^*}$ means that $d_{P_i(t)\cup P_j(t)}(m, b^*) < d_{P_i(t)\cup P_j(t)}(m, a^*)$. This implies that
\begin{align*}
d_{P_i\cup P_j}(x, b^*)&\leq d_{P_i\cup P_j}(m,b^*)+d_{P_i \cup P_j}(x,m)\\
&< d_{P_i\cup P_j}(m,a^*)+d_{P_i\cup P_j}(x,m)\\
&= d_{P_i \cup P_j}(x,a^*).
\end{align*}
This is a contradiction for $x\in W_{a^*}$. Similar considerations hold for $W_{b^*}$.
\end{IEEEproof}

The rest of this section is dedicated to proving convergence.  
Our first step is to show that the evolution determined by the 
Discrete Gossip Coverage Algorithm can be seen as a set-valued map. 
To this end, for any pair of robots $(i,j)\in\until{N}^2$, $i\not=j$, we define the map
$\map{T_{ij}}{ \ConnPart}{ \ConnPart}$ by
\begin{align*}
  T_{ij}(P) =
    (P_1,\dots,\widehat{P}_i,\dots,\widehat{P}_j, \dots,P_N), 
\end{align*}
where $\widehat{P}_i = W_{a^*}$ and $\widehat{P}_j = W_{b^*}$.

If  at time $t\in \realnonnegative$ the pair $(i,j)$ and no other pair of robots perform an iteration of the Pairwise Partitioning Rule, then the dynamical system on the space of partitions is described by
\begin{equation}\label{eq:Tij}
  P(t^+)=T_{ij}\left( P(t) \right).
\end{equation}
We define the set-valued map $\setmap{T}{\ConnPart}{\ConnPart}$ as
\begin{equation}\label{eq:AlgoT}
T(P)=\setdef{T_{ij}(P)}{(i,j)\in\until{N}^2, i\not=j}.
\end{equation}
Observe that~\eqref{eq:Tij} can then be rewritten as $P(t^+)\in T(P(t))$.

The next two Propositions state facts whose validity is ensured by 
Lemma~\ref{lemma:OnMotionProtocol} of Appendix~\ref{sec:appendix_B} which states a key property of the Random Destination \& Wait Motion Protocol.

\begin{proposition}[Persistence of Exchanges]\label{prop:tk}
Consider~$N$ robots implementing the Discrete Gossip Coverage Algorithm. Then, there almost surely exists  an increasing sequence of time instants $\left\{t_k\right\}_{k \in \integernonnegative}$ such that
$
P(t_k^+)=T_{ij}(P(t_k))
$
for some $(i,j)\in \mathcal{E}(P(t_k))$. 
\end{proposition}
\begin{IEEEproof}
The proof follows directly from Lemma~\ref{lemma:OnMotionProtocol} which implies that the time between two consecutive pairwise communications is almost surely finite. 
\end{IEEEproof}

\vspace{0.1cm}

The existence of time sequence $\left\{t_k\right\}_{k \in \integernonnegative}$ allows us to to express the evolution generate by the Discrete Gossip Coverage Algorithm as a discrete time process.  Let $P(k):=P(t_k)$ and $P(k+1):=P(t_k^+)$, then 
$$
P(k+1)\in T\left( P(k) \right)
$$
where $\setmap{T}{\ConnPart}{\ConnPart}$ is defined as in~\eqref{eq:AlgoT}. 

Given $k \in  \integernonnegative$, let $\mathcal{I}_k$ denote the information which completely characterizes the state of Discrete Gossip Coverage Algorithm just after the $k$-th iteration of the partitioning rule, i.e., at time $t_{k-1}^+$. Specifically, $\mathcal{I}_k$ contains the information related to the partition $P(k)$, the positions of the robots at $t_{k-1}^+$, and whether each robot is in the \emph{waiting} or \emph{moving} state at $t_{k-1}^+$. 
The following result characterizes the probability that, given $\mathcal{I}_k$, the $(k+1)$-th iteration of the partitioning rule is governed by any of the maps $T_{ij}$, $(i,j)\in \mathcal{E}(P(k))$.

\begin{proposition}[Probability of Communication]\label{prop:pi}
Consider a team of $N$ robots with capacities (C1), (C2), (C3), and (C4) implementing the Discrete Gossip Coverage Algorithm. Then, there exists a real number $\bar{\pi} \in (0,1)$, such that, for any $k\in \integernonnegative$ and $(i,j)\in \mathcal{E}(P(k))$
\begin{equation*}
\Prob\left[P(k+1)=T_{ij}(P(k)) \;|\; \mathcal{I}_k\right]\geq \bar{\pi}.
\end{equation*}
\end{proposition}
\begin{IEEEproof}
Assume that at time $\bar t$ one pair of robots communicates. Given a pair $(\bar i,\bar j)\in \E(P(\bar t))$, we must find a lower bound for the probability that  $(\bar i,\bar j)$ is the communicating pair.
Since all the Poisson communication processes have the same intensity, the distribution of the chance of communication is uniform over the pairs which are ``able to communicate,'' i.e., closer than $\rcomm$ to each other. Thus, we must only show that $(\bar i,\bar j)$ has a positive probability of being able to communicate at time $\bar t$, which is equivalent to showing that $(\bar i,\bar j)$ is able to communicate for a positive fraction of time with positive probability. The proof of Lemma~\ref{lemma:OnMotionProtocol} implies that with probability at least $\alpha/(1-e^{-\subscr{\lambda}{comm}\tau})$ any pair in $\E(P(\bar t))$ is able to communicate for a fraction of time not smaller than $\frac{\tau}{\Delta},$ where $\alpha$ and $\Delta$ are defined in the proof of Lemma~\ref{lemma:OnMotionProtocol}. Hence the result follows.
\end{IEEEproof}
\vspace{0.1cm}

The property in Proposition~\ref{prop:pi} can also be formulated as follows. 
Let
$\map{\switchsig}{\integernonnegative}{\left\{(i,j)\in\until{N}^2, i\not=j\right\}}$ be the stochastic process such that $\switchsig(k)$ is the communicating pair at time $k$.
Then, the sequence of pairs of robots performing the partitioning rule at time instants $\left\{t_k\right\}_{k \in \integernonnegative}$ can be seen as a realization of the process $\sigma$, which satisfies 
\begin{equation}\label{eq:RanPer2}
\Prob\big[\switchsig(k+1)=(i,j) \;|\;  \switchsig(k)\big] \geq \bar{\pi}
\end{equation}
for all $(i,j)\in \mathcal{E}(P(k))$.
\smallskip

Next we show that the cost function decreases whenever the application of $T$ from \eqref{eq:AlgoT} changes the territory partition. This fact is a key ingredient to apply  Lemma~\ref{lem:finite-LaSalle}.

\begin{lemma}[Decreasing Cost Function]\label{lemma:Tdecr}

Let $P\in\ConnPart$ and let $P^+\in T(P)$. 
If $P^+ \neq P$,  then $\Hexp(P^+) < \Hexp(P)$.
\end{lemma}

\begin{IEEEproof}
Without loss of generality assume that $(i,j)$ is the pair executing the Pairwise Partitioning Rule. 
Then
\begin{align*}
&\Hexp(P^+)- \Hexp(P)\\
&\qquad\qquad\qquad=  \Hone(\Cd(P_i^+); P_i^+)+ \Hone(\Cd(P_j^+); P_j^+)\\
&\qquad\qquad\qquad\quad-\Hone(\Cd(P_i); P_i)- \Hone(\Cd(P_j); P_j).
\end{align*}
According to the definition of the Pairwise Partitioning Rule we have that if $P_i^+\neq P_i$, $P_j^+\neq P_j$, then 
\begin{align*}
&\Hone(\Cd(P_i^+); P_i^+)+ \Hone(\Cd(P_j^+); P_j^+)\\
&\qquad\qquad \leq \Hone(a^*; P_i^+)+ \Hone(b^*; P_j^+)\\
&\qquad\qquad < \Hone(\Cd(P_i); P_i)+ \Hone(\Cd(P_j); P_j)
\end{align*}
from which the statement follows.
\end{IEEEproof}

\smallskip
We now complete the proof of the main result, Theorem~\ref{th:main}. 

\begin{IEEEproof}[Proof of Theorem~\ref{th:main} statement (\ref{item:convergence})]
Note that the algorithm evolves in a finite space of partitions, and by Theorem~\ref{th:main} statement (\ref{item:well-posedness}), the set $\ConnPart$ is strongly positively
invariant. This fact implies that assumption~(i) of Lemma~\ref{lem:finite-LaSalle} is satisfied.
From Lemma~\ref{lemma:Tdecr} it follows that assumption (ii) is also satisfied, with $\Hexp$ playing the role of the function $U$.  
Finally, the property in~\eqref{eq:RanPer2} is equivalent to the property of \emph{persistent random switches} stated in Assumption~(iii) of Lemma~\ref{lem:finite-LaSalle}, for the special case $h=1$. 
Hence, we are in the position to apply Lemma~\ref{lem:finite-LaSalle} and conclude convergence in finite-time to an element of the intersection of the equilibria of the maps $T_{ij}$, which by definition is the set of the pairwise-optimal partitions.
\end{IEEEproof}

\section{Experimental Methods \& Results}
\label{sec:results}

To demonstrate the utility and study practical issues of the Discrete Gossip Coverage Algorithm, we implemented it using the open-source Player/Stage robot control system \cite{BG-RTV-AH:03} and the Boost Graph Library (BGL) \cite{JBS-LQL-AL:07}.  All results presented here were generated using Player 2.1.1, Stage 2.1.1, and BGL 1.34.1.  To compute distances in uniform edge weight graphs we extended the BGL breadth-first search routine with a distance recorder event visitor.

\begin{figure*}[t]
\centering
\subfigure
{
    \includegraphics[width=0.98\columnwidth]{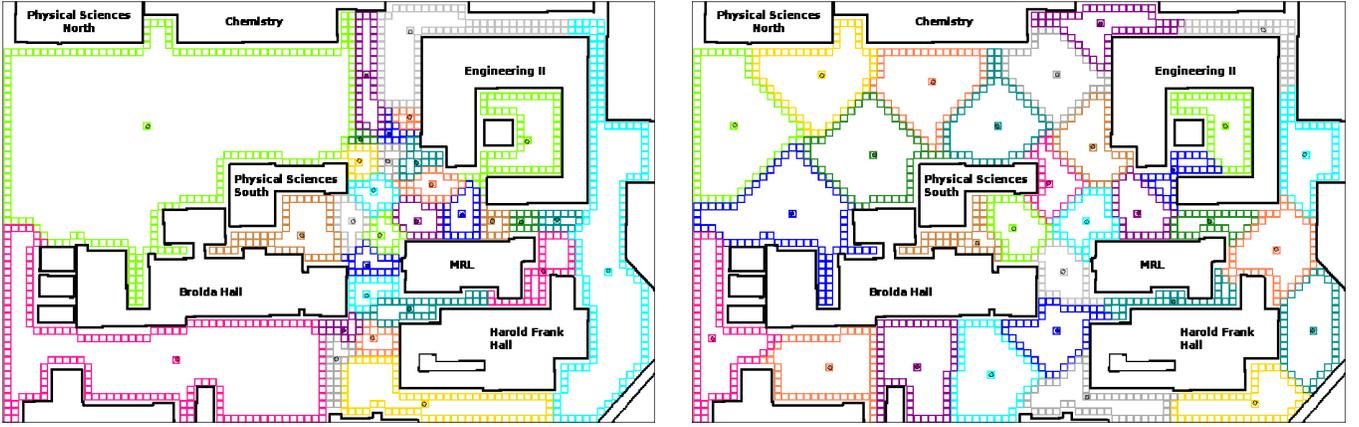}
}
\hspace{0.0in}
\subfigure
{
    \includegraphics[width=0.98\columnwidth]{sim_campus_2.eps}
}
\caption{Images of starting and final partitions for a simulation with 30 robots providing coverage of a portion of campus at UCSB.}
\label{fig:large_sim}
\end{figure*}

\subsection{Large-scale Simulation}

To evaluate the performance of our gossip coverage algorithm with larger teams, we tested 30 simulated robots partitioning a map representing a $350m \times 225m$ portion of campus at the University of California at Santa Barbara.  As shown in Fig.~\ref{fig:large_sim}, the robots are tasked with providing coverage of the open space around some of the buildings on campus, a space which includes a couple open quads, some narrower passages between buildings, and a few dead-end spurs.  
For this large environment the simulated robots are $2m$ on a side and can move at $3.0\tfrac{m}{s}$.  Each territory cell is $3m \times 3m$.

In this simulation we handle communication and partitioning as follows.  The communication range is set to $30m$ (10 edges in the graph) with $\subscr{\lambda}{comm} = 0.3\frac{\text{comm}}{s}$. 
The robots wait at their destination vertices for $\tau = 3.5s$.  This value for $\tau$ was chosen so that on average one quarter of the robots are waiting at any moment.  Lower values of $\tau$ mean the robots are moving more of the time and as a result more frequently miss connections, while for higher $\tau$ the robots spend more time stationary which also reduces the rate of convergence.
With the goal of improving communication, we implemented a minor modification to the motion protocol:  each robot picks its random destination from the cells forming the open boundary\footnote{The open boundary of $P_i$ is the set of vertices in $P_i$ which are adjacent to at least one vertex owned by another agent.} of its territory.  
In our implementation, the full partitioning loop may take $5$ seconds for the largest initial territories in Fig.~\ref{fig:large_sim}.  We chose to stop the loop after a quarter second for this simulation to verify the anytime computation claim.

The 30 robots start clustered in the center of the map between Engineering II and Broida Hall, and an initial Voronoi partition is generated from these starting positions.
This initial partition is shown on the left in Fig.~\ref{fig:large_sim} with the robots positioned at the  centroids of their starting regions.  The initial partition has a cost of $37.1m$.
The team spends about 27 minutes moving and communicating according to the Discrete Gossip Coverage Algorithm before settling on the final partition on the right of Fig.~\ref{fig:large_sim}.  The coverage cost of the final equilibrium improved by $54\%$ to $17.1m$.  Visually, the final partition is also dramatically more uniform than the initial condition.
This result demonstrates that the algorithm is effective for large teams in large non-convex environments.

\begin{figure}[t]
\centering
\psfrag{x_label}{\small{Time (minutes)}}
\psfrag{y_label}{\small{Total cost $\Hexp$}}
\includegraphics[height=3.32cm]{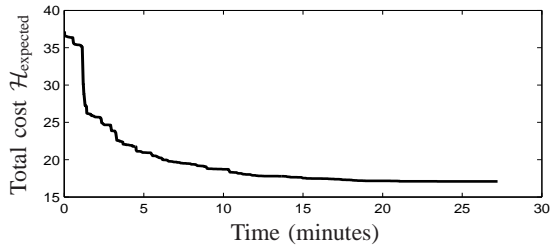}
\caption{Graph of the cost $\Hexp$ over time for the simulation in Fig.~\ref{fig:large_sim}.}
\label{fig:large_sim_cost}
\end{figure}

Fig.~\ref{fig:large_sim_cost} shows the evolution of $\Hexp$ during the simulation.  
The largest cost improvements happen early when the robots that own the large territories on the left and right of the map communicate with neighbors with much smaller territories.  These big territory changes then propagate through the network as the robots meet and are pushed and pulled towards a lower cost partition.

\subsection{Implementation Details}
\label{sec:implementation}

\begin{figure*}[t]
\centering
\begin{minipage}{0.325\linewidth}  
  \includegraphics[width=\linewidth]{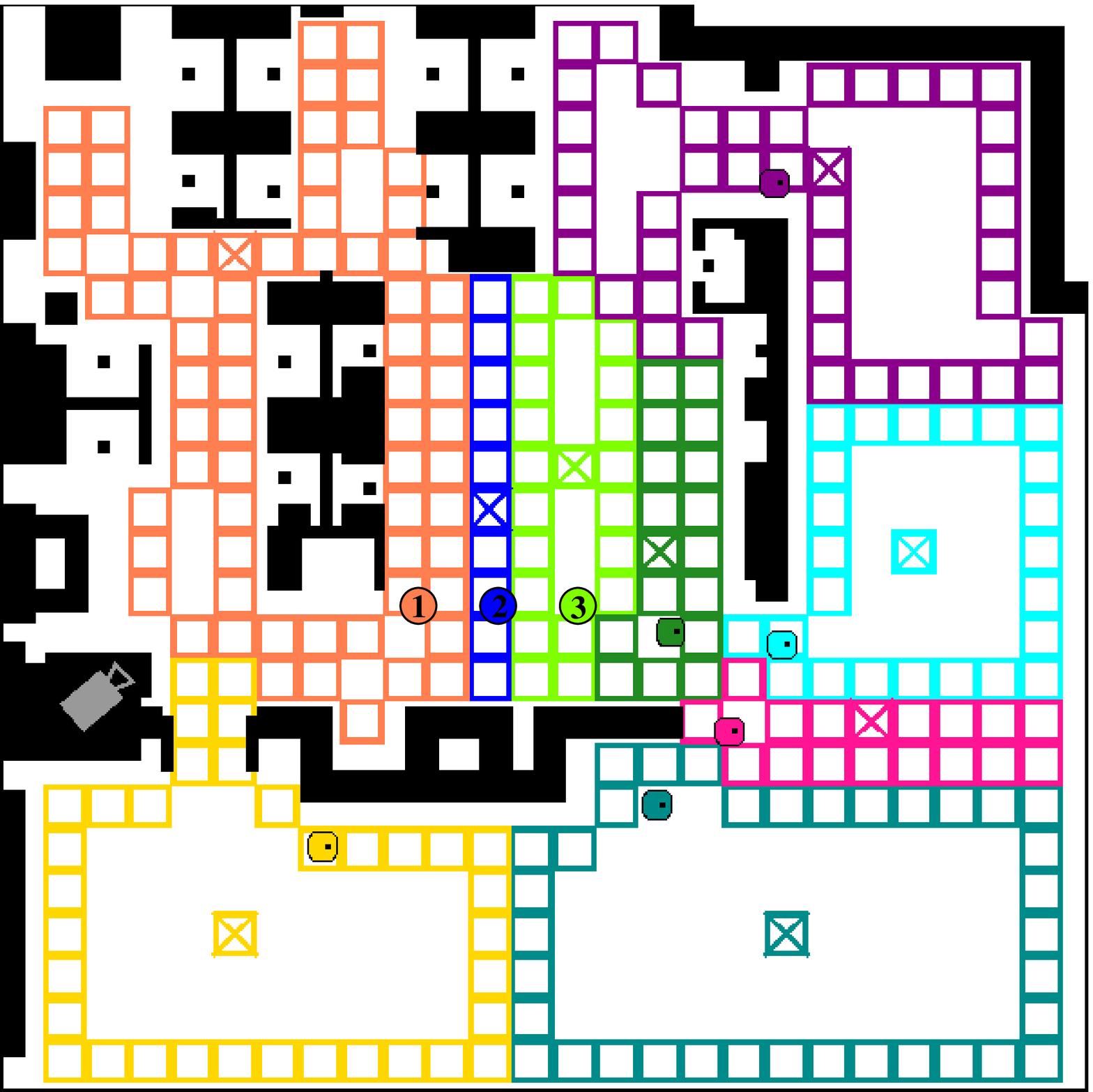}
  
  \vspace{0.03in}
  
  \includegraphics[width=\linewidth]{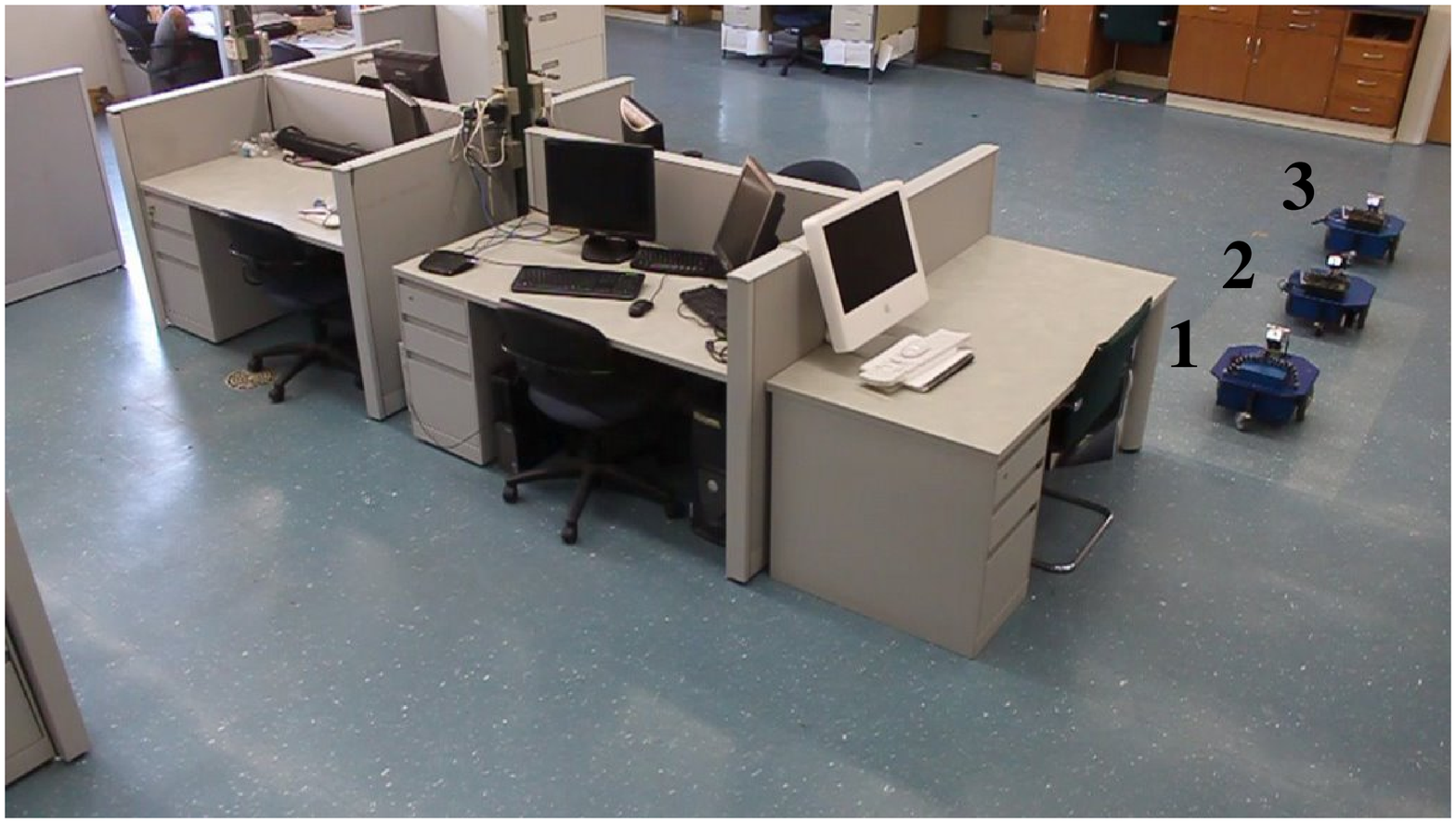}
\end{minipage}
\begin{minipage}{0.325\linewidth}  
  \includegraphics[width=\linewidth]{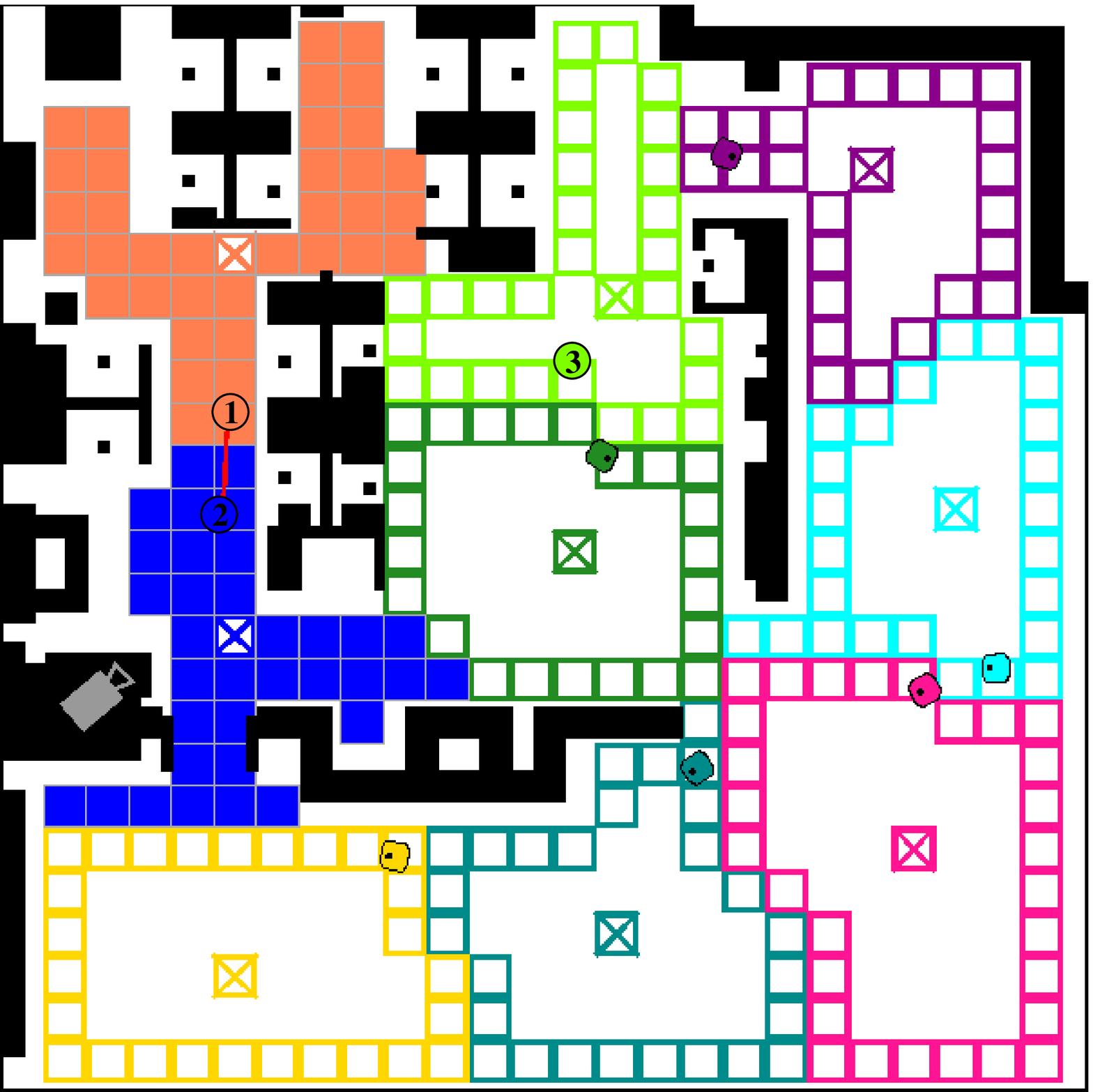}
  
  \vspace{0.03in}
  
  \includegraphics[width=\linewidth]{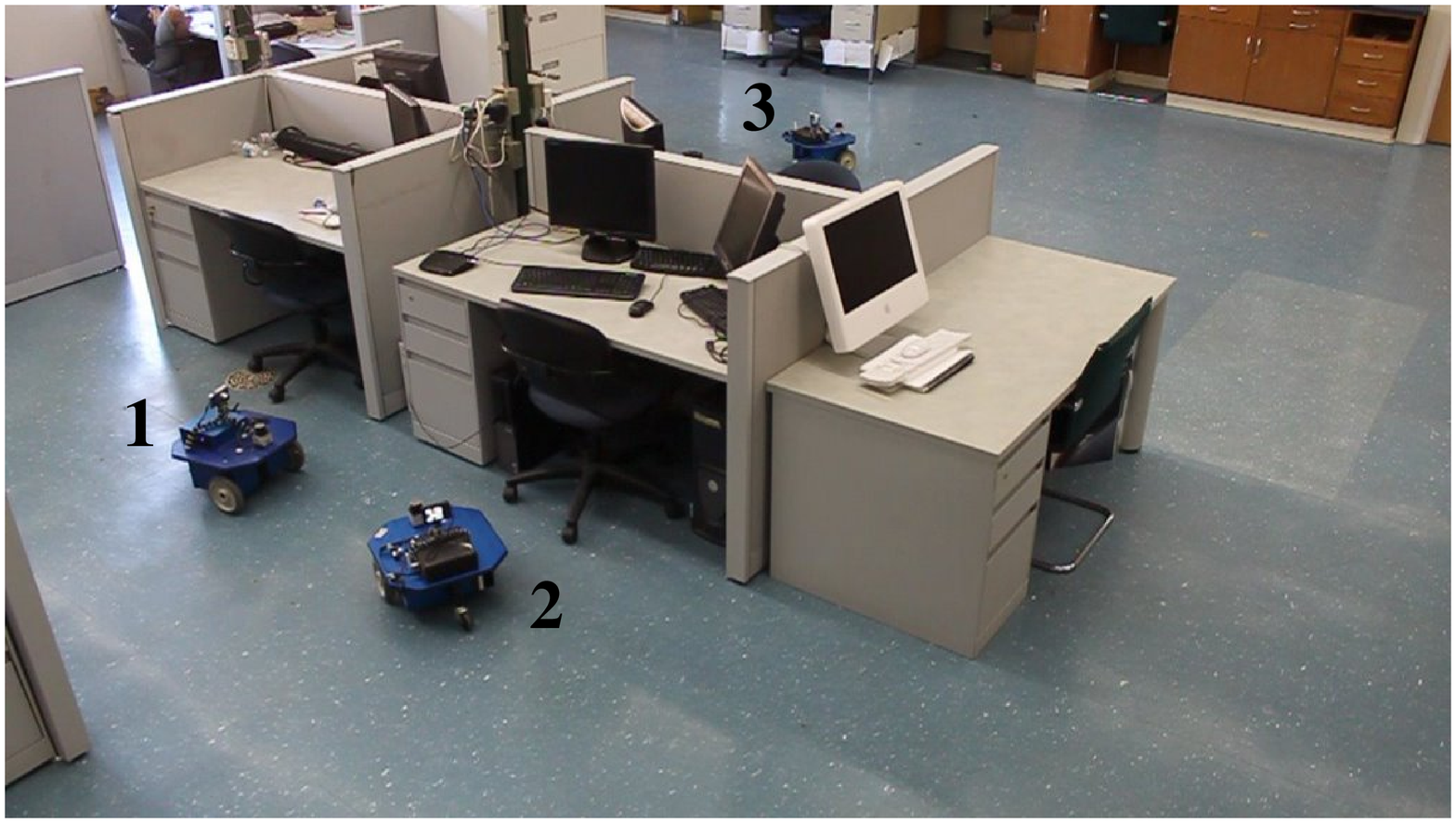}
\end{minipage}
\begin{minipage}{0.325\linewidth}  
  \includegraphics[width=\linewidth]{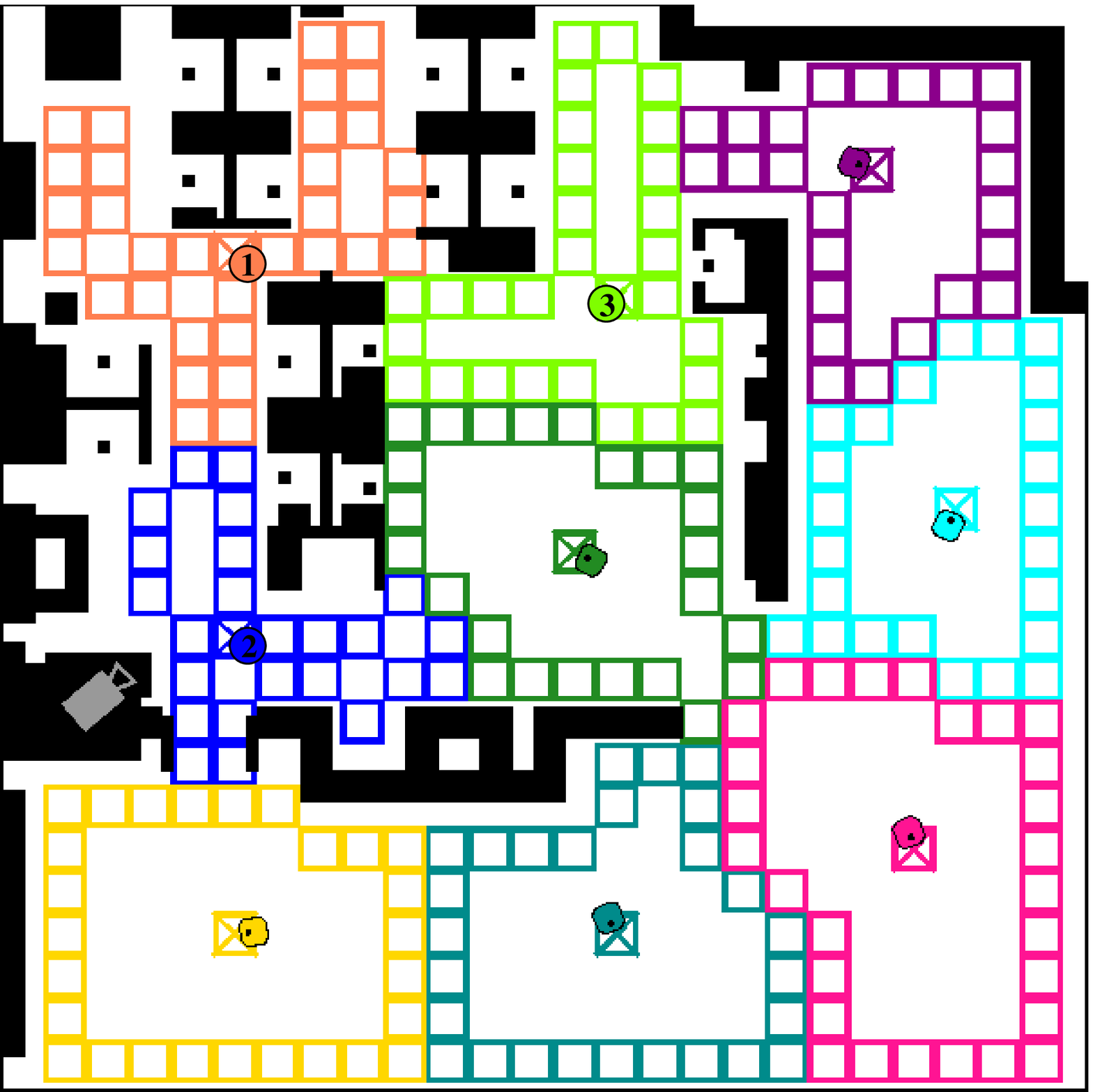}
  
  \vspace{0.03in}
  
  \includegraphics[width=\linewidth]{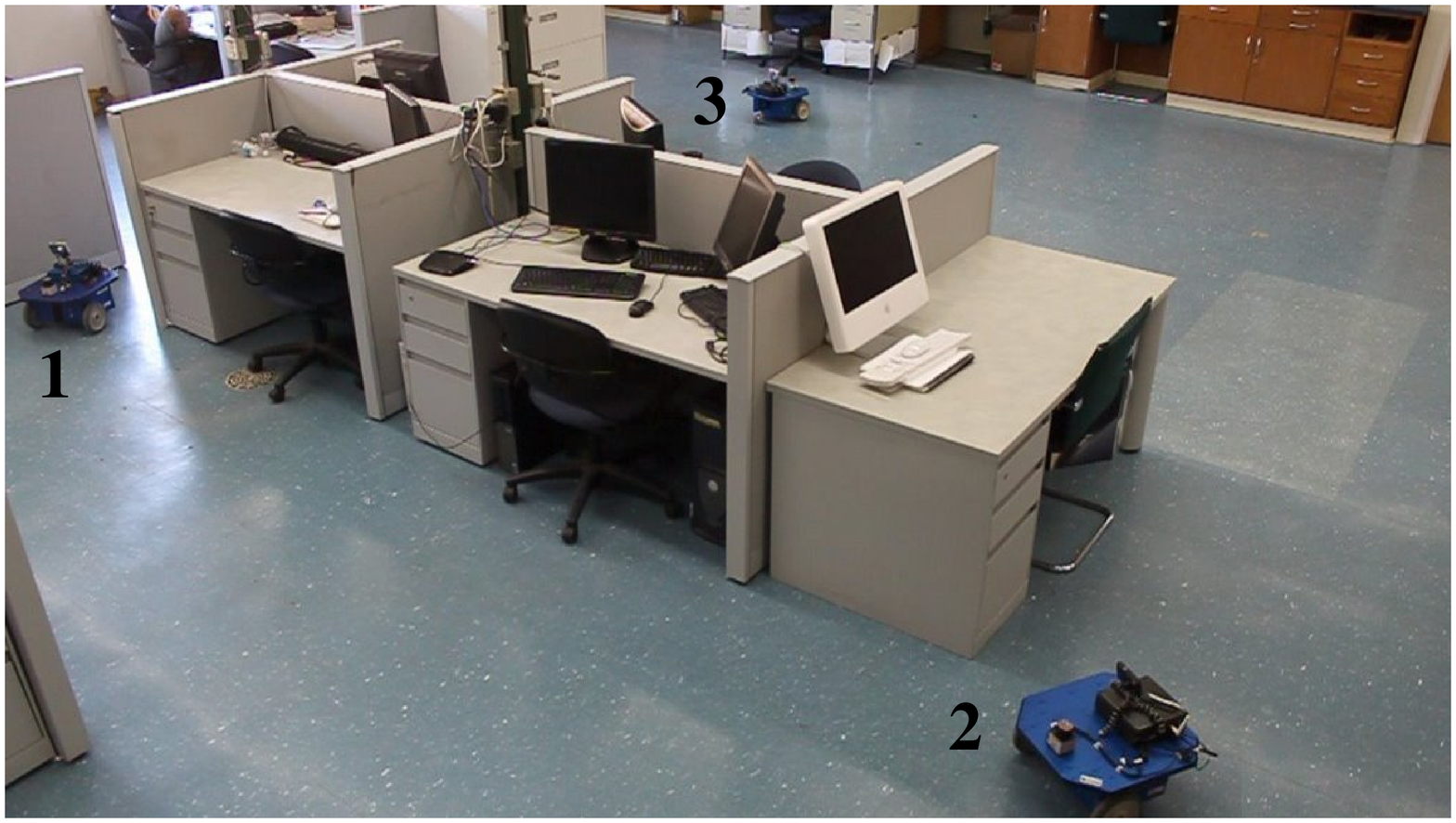}
\end{minipage}
\caption{Each column contains a territory map and the
  corresponding overhead camera image for a step of the
  hardware-in-the-loop simulation.  The position of the camera in the
  environment is shown with a camera icon in the territory map.  The
  physical robots are numbered 1, 2, and 3 and have the orange, blue, and lime green partitions.  Their
  positions in each territory map are
  indicated with numbered circles.}
\label{fig:experiment}
\end{figure*}

We conducted an experiment to test the algorithm using three physical robots in our lab, augmented by six simulated robots in a synthetic environment extending beyond the lab.  
Our lab space is $11.3m$ on a side and is represented by the upper left portion of the territory maps in Fig.~\ref{fig:experiment}.  The territory graph loops around a center island of desks.  We extended the lab space through three connections into a simulated environment around the lab, producing a $15.9m \times 15.9m$ environment.
The map of the environment was specified with a $0.15m$ bitmap which we overlayed with a $0.6m$ resolution occupancy grid representing the free territory for the robots to cover.  The result is a lattice-like graph with all edge weights equal to $0.6m$.  The $0.6m$ resolution was chosen so that our physical robots would fit easily inside a cell.

Additional details of our implementation are as follows.

\subsubsection*{Robot hardware}

\begin{figure}[t]
\centering
\psfrag{rangefinder}{\small Rangefinder}
\psfrag{drive wheel}{\small Drive wheel}
\psfrag{Computer}{\small Computer}
\psfrag{Rear caster}{\small Rear caster}
\includegraphics[height=3.32cm]{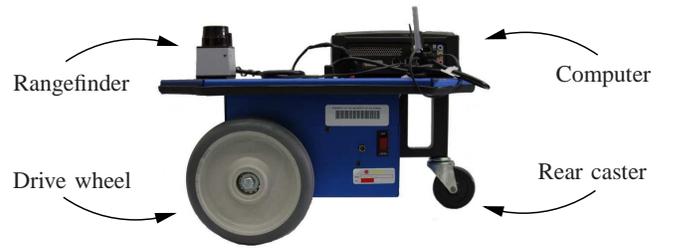}
\caption{Erratic mobile robot with URG-04LX laser rangefinder.}
\label{fig:robot}
\end{figure}

We use Erratic mobile robots from Videre Design, as shown in Fig.~\ref{fig:robot}.
The vehicle platform has a roughly square footprint $(40 cm \times 37cm)$, with two differential
drive wheels and a single rear caster.
Each robot carries an onboard computer with a 1.8Ghz Core 2 Duo processor, 1 GB
of memory, and 802.11g wireless communication.  
For navigation and localization, each robot is equipped with a Hokuyo URG-04LX laser rangefinder.
The rangefinder scans $683$ points over $240\degree$ at $10Hz$ with a range of $5.6$ meters.

\subsubsection*{Experiment setup}

Our mixed physical and virtual robot experiments are run from a central computer which is
attached to a wireless router so it can communicate with the physical robots.  
The central computer creates a simulated world using Stage which mirrors and extends the 
real space in which the physical robots operate.  The central computer also simulates the virtual
members of the robot team.  These virtual robots are modeled off of our hardware: they are differential drive
with the same geometry as the Erratic platform and use simulated Hokuyo URG-04LX rangefinders.

\subsubsection*{Localization}

We use the \texttt{amcl} driver in Player which implements Adaptive Monte-Carlo Localization~\cite{DF-WB-FD-ST:01}.  The physical robots are provided with a map of our lab with a $15cm$ resolution and told their starting pose within the map.  We set an initial pose standard deviation of $0.9m$ in position and $12\degree$ in orientation, and request localization updates using $50$ of the sensor's range measurements for each change of $2cm$ in position or $2\degree$ in orientation reported by the robot's odometry system.  We then use the most likely pose estimate output by \texttt{amcl} as the location of the robot.  For simplicity and reduced computational demand, we allow the virtual robots access to perfect localization information.

\subsubsection*{Motion Protocol}

Each robot continuously executes the Random Destination \& Wait Motion Protocol, with navigation handled by the \texttt{snd} driver in Player which implements Smooth Nearness Diagram navigation~\cite{JWD-FB:08a}.
For \texttt{snd} we set the robot radius parameter to $22cm$, obstacle avoidance distance to $0.7m$, and maximum speeds
to $0.4 \frac{m}{s}$ and $40 \tfrac{\degree}{s}$. 
The \texttt{snd} driver
is a local obstacle avoidance planner, so we feed it a series of waypoints every couple meters
along paths found in $\G$.  
We consider a robot to have achieved its target location when it is within $20cm$ and it will then wait for $\tau = 3.5s$.
For the physical robots the motion protocol and navigation processes run on board, while 
there are separate threads for each virtual robot on the central computer.

\subsubsection*{Communication and Partitioning}

As the robots move, a central process monitors their positions and simulates the range-limited gossip communication model between both real and virtual robots.  
We set $r_{comm} = 2.5m$ and $\lambda_{comm} = 0.3\frac{\text{comm}}{s}$.  These parameters were chosen so that the robots would be likely to communicate when separated by at most four edges, but would also sometimes not connect despite being close.
When this process determines two robots
should communicate, it informs the robots who then perform the
Pairwise Partitioning Rule.  Our pairwise communication implementation is blocking: if robot $i$ is
exchanging territory with $j$, then it informs the match making process
that it is unavailable until the exchange is complete.

\subsection{Hardware-in-the-Loop Simulation}

\begin{figure}[t]
\centering
\psfrag{x_label}{Time (minutes)}
\psfrag{y_label_1}{Robot costs (m)}
\psfrag{y_label_2}{$\Hexp$ (m)}
\includegraphics[width=0.99\columnwidth]{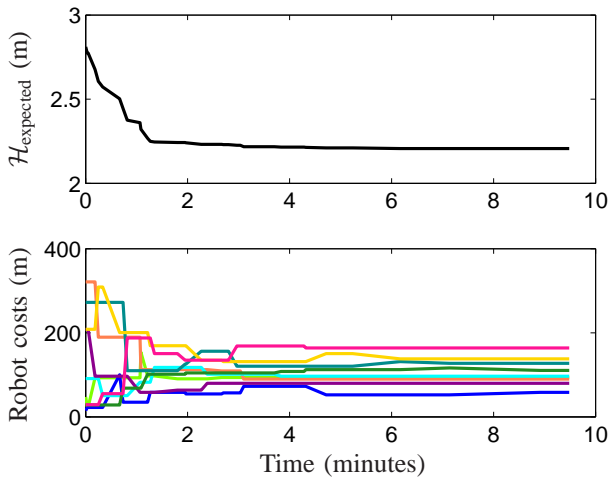}
\caption{Evolution of cost functions during the experiment in Fig.~\ref{fig:experiment}. The total cost $\Hexp$ is shown above in black, while $\Hone$ for each robot is shown below in the robot's color.}
\label{fig:exp_cost}
\end{figure}

The results of our experiment with three physical robots and six simulated robots are shown in Figs.~\ref{fig:experiment} and \ref{fig:exp_cost}.  The left column in Fig.~\ref{fig:experiment} shows the starting positions of the team of robots, with the physical robots, labeled 1, 2, and 3, lined up in a corner of the lab and the simulated robots arrayed around them.  The starting positions are used to generate the initial Voronoi partition of the environment.  The physical robots own the orange, blue, and lime green territories in the upper left quadrant.  We chose this initial configuration to have a high coverage cost, while ensuring that the physical robots will remain in the lab as the partition evolves.

In the middle column, robots 1 and 2 have met along their shared border and are exchanging territory.  In the territory map, the solid red line indicates 1 and 2 are communicating and their updated territories are drawn with solid orange and blue, respectively.  The camera view confirms that the two robots have met on the near side of the center island of desks.

The final partition at right in Fig.~\ref{fig:experiment} is reached after $9 \frac{1}{2}$ minutes.  All of the robots are positioned at the centroids of their final territories.  The three physical robots have gone from a cluster in one corner of the lab to a more even spread around the space.
Fig.~\ref{fig:exp_cost} shows the evolution of the cost function $\Hexp$ as the experiment progresses, including the costs for each robot.  As expected, the total cost never increases and the disparity of costs for the individual robots shrinks over time until settling at a pairwise-optimal partition.

In this experiment the hardware challenges of sensor noise, navigation, and uncertainty in position were efficiently handled by the \texttt{amcl} and \texttt{snd} drivers.  The coverage algorithm assumed the role of a higher-level planner, taking in position data from \texttt{amcl} and directing \texttt{snd}.  By far the most computationally demanding component was \texttt{amcl}, but the position hypotheses from \texttt{amcl} are actually unnecessary: our coverage algorithm only requires knowledge of the vertex a robot occupies.  If a less intensive localization method is available, the algorithm could run on robots with significantly lower compute power.

\subsection{Comparative analysis}
\label{sec:analysis}

In this subsection we present a numerical comparison of the performance of the Discrete Gossip Coverage Algorithm and the following two Lloyd-type algorithms.

\subsubsection*{Decentralized Lloyd Algorithm}
This method is from \cite{JC-SM-TK-FB:02j} and \cite{FB-JC-SM:09}, we describe it here for convenience.
At each discrete time instant $t \in \integersnonnegative$, each robot $i$ performs the following tasks: (1) $i$ transmits its position and receives the positions of all adjacent robots; (2) $i$ computes its Voronoi region $P_i$ based on the information received; and (3) $i$ moves to $\Cd(P_i)$.

\subsubsection*{Gossip Lloyd Algorithm}
This method is from \cite{JWD-RC-PF-FB:08z}.
It is a gossip algorithm, and so we have used the same communication model and the Random Destination \& Wait Motion Protocol to create meetings between robots.  Say robots $i$ and $j$ meet at time $t$, then the pairwise Lloyd partitioning rule works as follows: (1) robot $i$ transmits $P_i(t)$ to $j$ and vice versa; (2) both robots determine $U = P_i(t) \cup P_j(t)$; (3) robot $i$ sets $P_i(t^+)$ to be its Voronoi region of $U$ based on $\Cd(P_i(t))$ and $\Cd(P_j(t))$, and $j$ does the equivalent.

For both Lloyd algorithms we use the same tie breaking rule when creating Voronoi regions as is present in the Pairwise Partitioning Rule: ties go to the robot with the lowest index.

\begin{figure}[t]
\centering
\includegraphics[width=0.75\columnwidth]{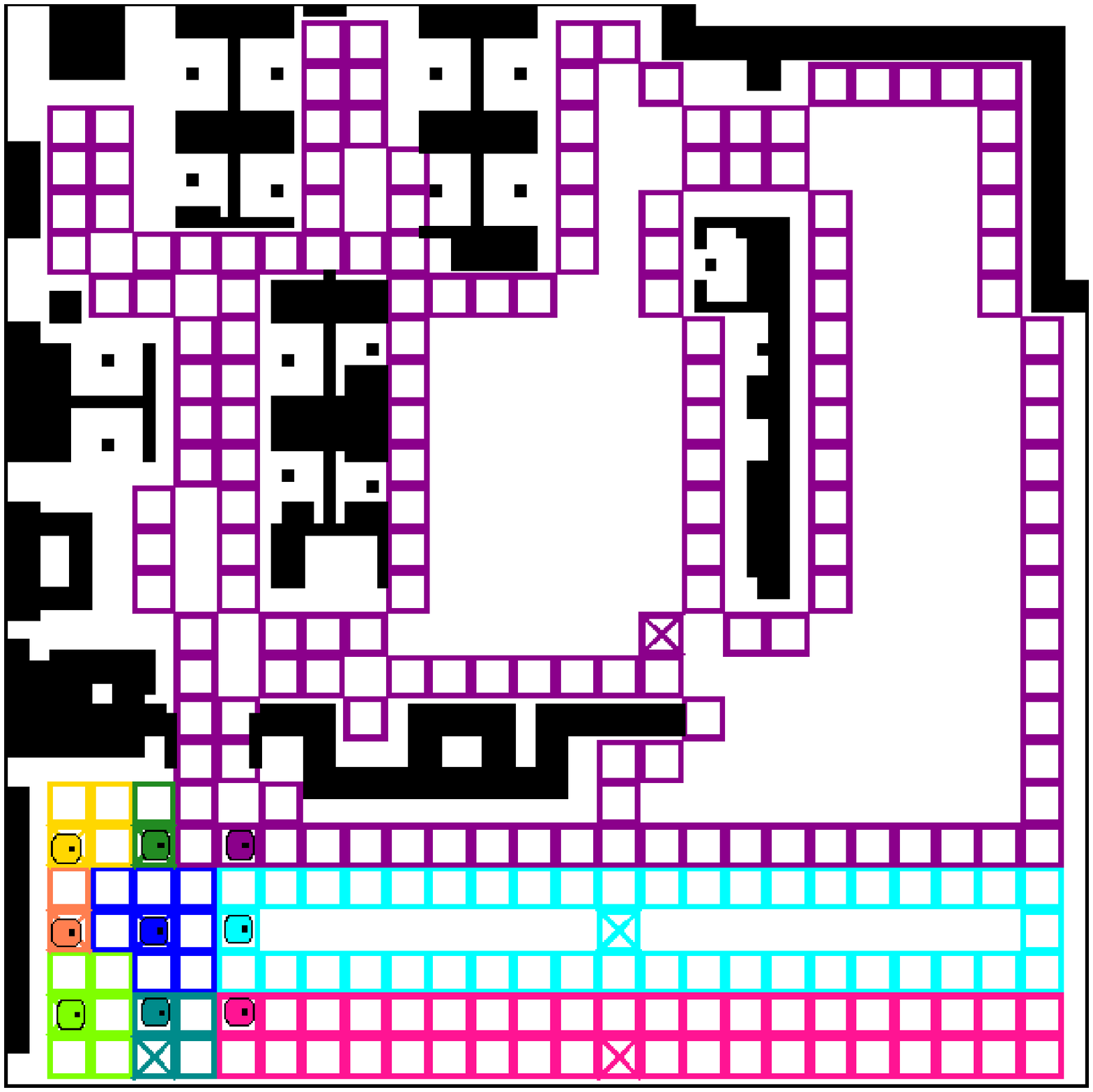}

\vspace{0.1cm}

\psfrag{xlabel}{Final cost (m)}
\psfrag{ylabel}{Simulation count}
\includegraphics[width=0.99\columnwidth]{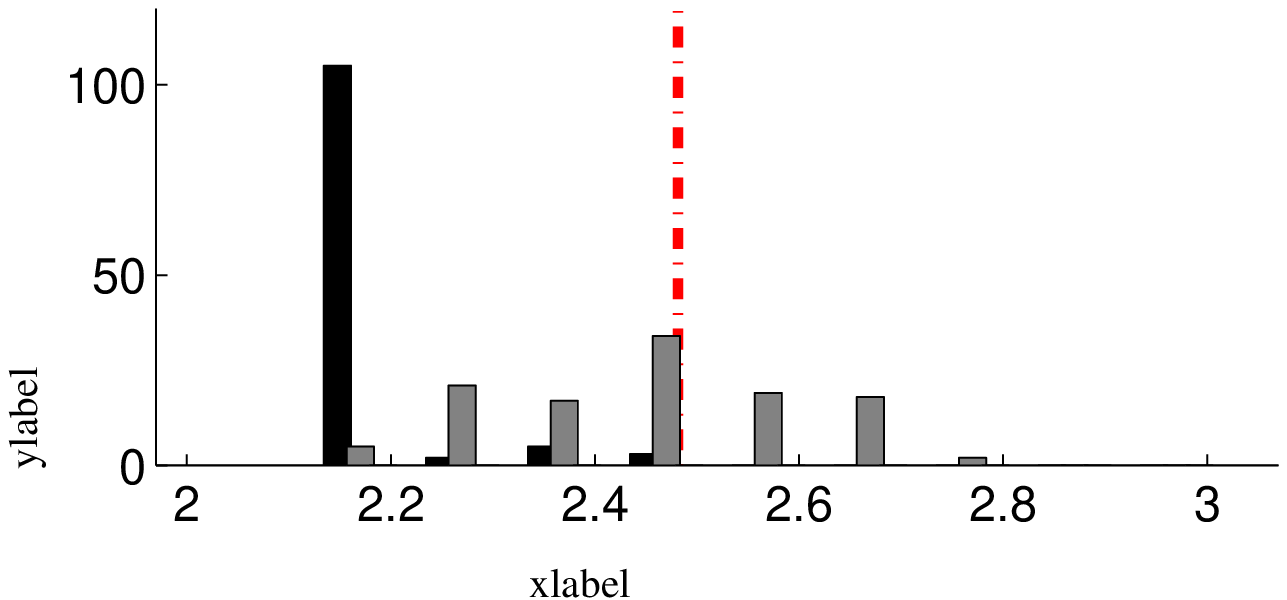}
\caption{Initial partition and histogram of final costs for a Monte Carlo test comparing the Discrete Gossip Coverage Algorithm (black bars), Gossip Lloyd Algorithm (gray bars), and Decentralized Lloyd Algorithm (red dashed line).  For the gossip algorithms, 116 simulations were performed with different sequences of pairwise communications.  The Decentralized Lloyd Algorithm is deterministic given an initial condition so only one final cost is shown.}
\label{fig:bad_start}
\end{figure}

Our first numerical result uses a Monte Carlo probability estimation
method from~\cite{RT-GC-FD:05} to place probabilistic bounds on the
performance of the two gossip algorithms.  Recall that the Chernoff bound
describes the minimum number of random samples $K$ required to reach a
certain level of accuracy in a probability estimate from independent
Bernoulli tests.  For an accuracy $\epsilon \in (0,1)$ and confidence $1
- \eta \in (0,1)$, the number of samples is given by
$K \geq \tfrac{1}{2\epsilon^2} \log \tfrac{2}{\eta}.$
For $\eta = 0.01$ and $\epsilon = 0.1$, at least 116 samples are required.

Figure~\ref{fig:bad_start} shows both the initial territory partition of the extended laboratory environment used and also a histogram of the final results for the following Monte Carlo test.  The environment and robot motion models used are described in Section~\ref{sec:implementation}.
Starting from the indicated initial condition, we ran 116 simulations of both gossip algorithms.  The randomness in the test comes from the sequence of pairwise communications.  These sequences were generated using: 
(1) the Random Destination \& Wait Motion Protocol with $q_i$ sampled uniformly from the open boundary of $P_i$ and $\tau = 3.5s$; and
(2) the range-limited gossip communication model with $r_{comm} = 2.5m$ and $\lambda_{comm} = 0.3\frac{\text{comm}}{s}$.

The cost of the initial partition in Fig.~\ref{fig:bad_start} is $5.48m$, while the best known partition for this environment has a cost of just under $2.18m$.  The histogram in Fig.~\ref{fig:bad_start} shows the final equilibrium costs for 116 simulations of the Discrete Gossip Coverage Algorithm (black) and the Gossip Lloyd Algorithm (gray).  It also shows the final cost using the Decentralized Lloyd Algorithm (red dashed line), which is deterministic from a given initial condition.
The histogram bins have a width of $0.10m$ and start from $2.17m$.  For the Discrete Gossip Coverage Algorithm, $105$ out of $116$ trials reach the bin containing the best known partition and the mean final cost is $2.23m$. The Gossip Lloyd Algorithm reaches the lowest bin in only $5$ of $116$ trials and has a mean final cost of $2.51m$.  The Decentralized Lloyd Algorithm settles at $2.48m$.  Our new gossip algorithm requires an average of $96$ pairwise communications to reach an equilibrium, whereas gossip Lloyd requires $126$.

Based on these results, we can conclude with $99\%$ confidence that there is at least an $80\%$ probability that 9 robots executing the Discrete Gossip Coverage Algorithm starting from the initial partition shown in Fig.~\ref{fig:bad_start} will reach a pairwise-optimal partition which has a cost within $4\%$ of the best known cost.  We can further conclude with $99\%$ confidence that the Gossip Lloyd Algorithm will settle more than $4\%$ above the best known cost at least $86\%$ of the time starting from this initial condition.

\begin{figure}[t]
\centering
\psfrag{xlabel}{Final cost (m)}
\psfrag{ylabel}{Simulation count}
\includegraphics[width=0.99\columnwidth]{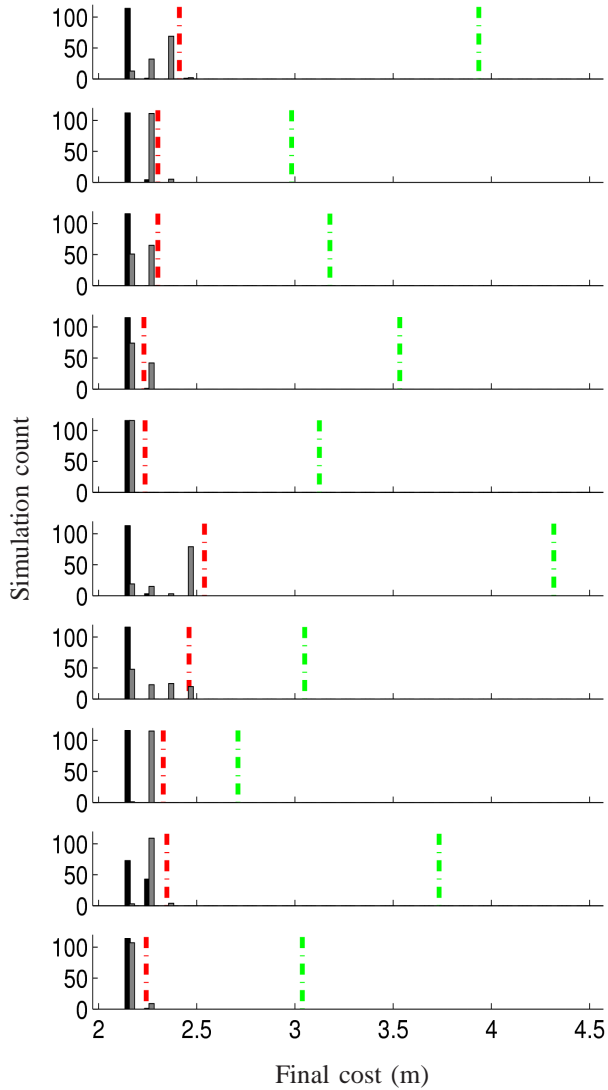}
\caption{Histograms of final costs from 10 Monte Carlo tests using random initial conditions in the environment shown in Fig.~\ref{fig:bad_start} comparing Discrete Gossip Coverage Algorithm (black bars), Gossip Lloyd Algorithm (gray bars), and Decentralized Lloyd Algorithm (red dashed line).  For the gossip algorithms, 116 simulations were performed with different sequences of pairwise communications.  The Decentralized Lloyd Algorithm is deterministic given an initial condition so only one final cost is shown.  The initial cost for each test is drawn with the green dashed line.}
\label{fig:multi_compare}
\end{figure}

Figure~\ref{fig:multi_compare} compares final cost
histograms for $10$ different initial conditions for the same environment
and parameters as described above.
Each initial condition was created by selecting unique starting locations
for the robots uniformly at random and using these locations to generate
an initial Voronoi partition.  The initial cost for each test is shown with
the green dashed line.  In 9 out of 10 tests the Discrete Gossip Coverage Algorithm reaches the histogram bin with the best known partition in at least $112$ of $116$ trials.
The two Lloyd methods get stuck in sub-optimal centroidal Voronoi partitions more than $4\%$ away from the best known partition in more than half the trials in 7 of 10 tests.

\section{Conclusion}
\label{sec:conclusion}

We have presented a novel distributed partitioning and coverage control algorithm which requires
only unreliable short-range communication between pairs of robots and works in non-convex environments.
The classic Lloyd approach to coverage optimization involves iteration
of separate centering and Voronoi partitioning steps.
For gossip algorithms, however, this separation 
is unnecessary computationally and we have
shown that improved performance can be achieved without it.  
Our new Discrete Gossip Coverage Algorithm provably
converges to a subset of the set of centroidal Voronoi partitions
which we labeled pairwise-optimal partitions.
Through numerical comparisons we demonstrated that this new subset of solutions
avoids many of the local minima in which Lloyd-type algorithms
can get stuck.

Our vision is that this partitioning and coverage algorithm will form the foundation of a distributed task servicing setup for teams of mobile robots.  The robots would split their time between servicing tasks in their territory and moving to contact their neighbors and improve the coverage of the space.  Our convergence results only require sporadic improvements to the cost function, affording flexibility in robot behaviors and capacities, and offering the ability to handle heterogeneous robotic networks.  In the bigger picture, this paper demonstrates the potential of gossip communication in distributed coordination algorithms.  There appear to be many other problems where this realistic and minimal communication model could be fruitfully applied.

\begin{appendices}
\section{}
\label{sec:appendix_A}

For completeness we present a convergence result for set-valued algorithms on finite state spaces, which can be recovered as a direct consequence of~\cite[Theorem~4.5]{FB-RC-PF:08u-web}.

Given a set  $X$, a set-valued map $\setmap{T}{X}{X}$ is a map which associates to an element $x\in X$ a subset $Z\subset X.$ A set-valued map is non-empty if
$T(x)\neq \emptyset$ for all $x\in{X}$. 
Given a non-empty set-valued map $T$, an evolution of the dynamical system associated to $T$ is a sequence
$\seqdef{x_n}{n\in\integernonnegative}\subset X$ where
$x_{n+1}\in T(x_n)$ for all $n\in\integernonnegative.$
A set $W\subset X$ is
\emph{strongly positively invariant} for $T$ if $T(w)\subset{W}$ for all
$w\in{W}$.

%

\begin{lemma}[Persistent random switches imply convergence]
  \label{lem:finite-LaSalle}
  Let $(X,d)$ be a finite metric space.  Given a collection of maps
  $\map{T_1,\ldots, T_m}{X}{X}$, define the set-valued map
  $\setmap{T}{X}{X}$ by $T(x)=\left\{T_1(x),\ldots, T_m(x)\right\}$.  Given
  a stochastic process $\map{\switchsig}{\integernonnegative}{\until{m}}$, consider
  an evolution $\seqdef{x_n}{n\in\integernonnegative}$ of $T$ satisfying
  $
    x_{n+1} = T_{\switchsig(n)}(x_n).
  $
  Assume that:
  \begin{enumerate}
  \item there exists a set $W\subseteq X$ that is strongly positively
    invariant for $T$;
  \item there exists a function $\map{U}{W}{\real}$ such that $U(w')< U(w)$,
    for all $w\in W$ and $w'\in T(w)\setminus\{w\}$; and
  \item there exist $p\in{(0,1)}$ and $k\in\mathbb{N}$ such that, for all
    $i\in\until{m}$ and $n\in\integernonnegative$, there exists
    $h\in\until{k}$ such that
    $
      \Prob\big[\switchsig(n+h)=i \,|\,  \switchsig(n),\dots,\switchsig(1)\big] \geq p.
    $
  \end{enumerate}
  For $i\in \until{m}$, let $F_i$ be the set of fixed points of $T_i$
  in $W$, i.e.,  $F_i=\setdef{w\in W}{T_i(w)=w}$.
  If $x_0\in{W}$, then the evolution  $\seqdef{x_n}{n\in\integernonnegative}$ converges almost surely in finite time to an element of the set  $(F_1\intersection \cdots \intersection F_m)$, i.e.,
  there exists almost surely $\tau\in\mathbb{N}$ such that, for some $\bar{x} \in (F_1\intersection \cdots \intersection F_m)$, $ x_n=\bar{x}$ for $n\geq \tau.$
\end{lemma}

\section{}
\label{sec:appendix_B}

This Appendix proves a property of the Random Destination \& Wait Motion Protocol which is needed to show the persistence of pairwise exchanges.

\begin{lemma} \label{lemma:OnMotionProtocol}
Consider $N$ robots implementing the Discrete Gossip Coverage Algorithm starting from an arbitrary $P \in \ConnPart$. Consider $t\in \realnonnegative$ and let $P(t)$ denote the partition at time $t$. Assume that at time $t$ no two robots are communicating. Then, there exist $\Delta>0$ and $\alpha\in(0,1)$, independent of $P(t)$ and the positions and states of the robots at time $t$, such that, for every $(i,j)\in\E(P(t))$,
$\Prob\left[(i,j) \text{ communicate within}\,\, (t,t+\Delta) \right]\ge \alpha.$
\end{lemma}
\begin{IEEEproof}
To begin, we define two useful quantities. Let 
$\displaystyle\PseudoDiam(Q):= \max_{P\in \ConnPart}\max_{P_i\in P}\max_{h,k\in P_i} d_{P_i}(h,k)$
be a pseudo-diameter for $Q$, 
and then choose $\Delta:=2\frac{\PseudoDiam(Q)}{v} + 2\tau$.
We fix a pair $(i,j)\in\E(P)$, and pick adjacent vertices $a\in P_i$, $b\in P_j$. 

Our goal is to lower bound the probability that $i$ and $j$ will communicate within the interval $\left(t,t+\Delta \right)$. To do so we construct {\em one} sequence of events of positive probability which enables such communication.
Consider the following situation: $i$ is in the \emph{moving} state and needs time $t_i$ to reach its destination $q_i$, whereas robot $j$ is in the \emph{waiting} state at vertex $q_j$ and must wait there for time $\tau_j\leq \tau$. 
We denote by $t(a)$ (resp. $t(b)$) the time needed for $i$ (resp. $j$) to travel from $q_i$ (resp. $q_j$) to $a$ (resp. $b$). 
Let $E_{i}$ be the event such that $i$ performs the following actions in $(t,t+\Delta)$ without communicating with any robot $k \ne j$:
\begin{enumerate}
\item $i$ reaches $q_i$ and waits at $q_i$ for the duration $\tau$; and
\item $i$ chooses vertex $a$ as its next destination and then stays at $a$ for at least $\Delta-t(a)-t_i-\tau$.
\end{enumerate}
Let $E_{j}$ be the event such that $j$ performs the following actions in $(t,t+\Delta)$ without communicating with any $k \ne i$:
\begin{enumerate}
\item $j$ waits at $q_j$ for the duration $\tau_j$; and
\item $j$ chooses vertex $b$ as its next destination and then stays at $b$ for at least $\Delta-t(b)-\tau_j$.
\end{enumerate} 
Let $E_{ij}=E_i \cap E_j$.

\newcommand{\lambdacomm}{\subscr{\lambda}{comm}}
Next, we lower bound the probability that event $E_i$ occurs.  Recall the definition of $\lambda_{\text{comm}}$ from Sec.~\ref{sec:Model}.
Since a robot can have at most $N-1$ neighbors, the probability that (i) of $E_i$ happens is lower bounded by  
$
 e^{-\lambdacomm \tau N}.
$
For (ii), the probability that $i$ chooses $a$ is $1/\card{P_i}$, which is lower bounded by  $1/\card{Q}$. Then, in order to spend at least $(\Delta-t(a)-t_i-\tau)$ at $a$, $i$ must choose $a$ for $\lceil \frac{\Delta-t(a)-t_i-\tau}{\tau} \rceil$ consecutive times. Finally, the probability that during this interval $i$ will not communicate with any robot other than $j$ is lower bounded by
$
e^{-\lambdacomm \Delta (N-2)}.
$
The probability that (ii) occurs is thus lower bounded by
$
\left(1 / \card{Q}\right)^{\lceil \frac{\Delta}{\tau} \rceil} e^{-\lambdacomm \Delta N}.
$
Combining the bounds for (i) and (ii), it follows that
$$
\Prob[E_i]\geq \bigl(\tfrac{1}{\card{Q}}\bigr)^{\lceil \frac{\Delta}{\tau} \rceil} e^{-\lambdacomm (\Delta+\tau) N}.
$$ 
The same lower bound holds for $\Prob[E_j]$, meaning that
\begin{align*}
\Prob\left[E_{ij}\right]&=\Prob\left[E_{i}\right]\, \Prob\left[E_{j}\right]
\geq  \bigl(\tfrac{1}{\card{Q}}\bigr)^{2 \lceil \frac{\Delta}{\tau} \rceil} e^{-2 \lambdacomm (\Delta+\tau) N}.
\end{align*}

If event $E_{ij}$ occurs, then robots $i$ and $j$ will be at adjacent vertices for an amount of time during the interval $(t,t+\Delta)$ equal to
$
\min \left\{\Delta-t(a)-t_i-\tau, \Delta-t(b)-\tau_j\right\}.
$
Since $t(a)$ and $t(b)$ are no more than $\frac{\PseudoDiam(Q)}{v}$, we can conclude that $i$ and $j$ will be within $\rcomm$ for at least $\tau$.
Conditioned on $E_{ij}$ occurring, the probability that $i$ and $j$ communicate in $(t,t+\Delta)$ is lower bounded by $1-e^{-\lambdacomm \tau}$. A suitable choice for $\alpha$ from the statement of the Lemma is thus
$$
\alpha= \bigl(\tfrac{1}{|Q|}\bigr)^{2 \lceil \frac{\Delta}{\tau} \rceil} e^{-2 \lambdacomm (\Delta+\tau) N}   \left(1-e^{-\lambdacomm\tau}\right).
$$  
It can be shown that this also constitutes a lower bound for the other possible combinations of initial states: 
robot $i$ is \emph{waiting} and robot $j$ is \emph{moving};
robots $i$ and $j$ are both \emph{moving}; and
robots $i$ and $j$ are both \emph{waiting}.
\end{IEEEproof}

\section{}
\label{sec:appendix_C}

In this appendix we provide the proof of Proposition~\ref{prop:OptPair} which states that any pairwise-optimal partition is also a centroidal Voronoi partition. 

\begin{IEEEproof}[Proof of Proposition~\ref{prop:OptPair}]
To create a contradiction, assume that $P \in \ConnPart$ is a pairwise-optimal partition but not a centroidal Voronoi partition.  In other words, there exist components $P_i$ and $P_j$ in $P$ and an element $x$ of one component, say $x\in P_i$, such that 
\begin{equation}\label{eq:not_voronoi}
d_G\left(x, \Cd(P_i)\right)> d_{G}\left(x, \Cd(P_j)\right).
\end{equation}  
Choose $P_j$ such that for all $k \neq j$
\begin{equation}\label{eq:lowest_j}
d_G\left(x, \Cd(P_k)\right) \geq d_{G}\left(x, \Cd(P_j)\right).
\end{equation}

Let $\short{a}{b}{G}$ be a shortest path in $G$ connecting $a$ to $b$ and let $m \in \short{x}{\Cd(P_j)}{G}$ be the first element of the path starting from $\Cd(P_j)$ which is not in $P_j$.  Let $\ell$ be such that $m \in P_\ell$.

If $m = x$, then from \eqref{eq:not_voronoi} and the definition of $\short{x}{\Cd(P_j)}{G}$ we have that 
\begin{align*}
&d_{P_i}\left(x, \Cd(P_i)\right) \geq d_{G}\left(x, \Cd(P_i)\right)\\
&\qquad\qquad > d_{G}\left(x, \Cd(P_i)\right) = d_{P_i \cup P_j}\left(x, \Cd(P_j)\right)
\end{align*}
which, since $x \in P_i$, creates a contradiction of the fact that $P$ is pairwise-optimal.

If $m \neq x$, then, given \eqref{eq:lowest_j}, one of these two conditions holds:
\begin{enumerate}
\item $d_G\left(m, \Cd(P_\ell)\right) > d_{G}\left(m, \Cd(P_j)\right)$, or
\item $d_G\left(m, \Cd(P_\ell)\right) = d_{G}\left(m, \Cd(P_j)\right)$.
\end{enumerate}
In the first case, we again have a contradiction using the same logic above with $m$ in place of $x$.  In the second case, we must further consider whether there exists a $\short{m}{\Cd(P_\ell)}{G}$ such that every vertex in $\short{m}{\Cd(P_\ell)}{G}$ is also in $P_\ell$.  If there is not such a path, then 
$$d_{P_\ell}\left(m, \Cd(P_\ell)\right) > d_G\left(m, \Cd(P_\ell)\right) = d_{P_\ell \cup P_j}\left(m, \Cd(P_j)\right)$$
and we again have a contradiction as above.  If there is such a path, then we can instead repeat this analysis using using $\ell$ in place of $j$ and considering the path formed by this $\short{m}{\Cd(P_\ell)}{G}$ and the vertices in $\short{x}{\Cd(P_j)}{G}$ after $m$.  Since the next vertex playing the role of $m$ must be closer to $x$, we will eventually find a vertex which creates a contradiction.
\end{IEEEproof}

\end{appendices}

\bibliographystyle{ieeetr}

\end{document}